\pdfoutput=1

\documentclass[11pt]{article}

\usepackage{acl}

\usepackage{times}
\usepackage{latexsym}
\usepackage[T1]{fontenc}
\usepackage[utf8]{inputenc}
\usepackage{microtype}
\usepackage{inconsolata}

\usepackage{graphicx}
\usepackage{booktabs}
\usepackage{multirow}
\usepackage{balance}
\usepackage{soul}
\usepackage{enumitem}
\usepackage{chngcntr}  
\usepackage{tcolorbox}
\usepackage{listings}
\usepackage{subcaption}
\usepackage{tabularx}
\usepackage{fontawesome5}
\usepackage{float}
\usepackage[capitalize,nameinlink,noabbrev]{cleveref}
\usepackage{pifont} 
\usepackage{siunitx}
\usepackage{longtable}
\usepackage{amssymb}
\usepackage{marvosym} 
\newlist{todolist}{itemize}{2}
\setlist[todolist]{label=$\square$}

\newcommand\rurl[1]{%
  \href{https://#1}{\nolinkurl{#1}}%
}

\definecolor{pastelblue}{HTML}{A1C9F4}
\definecolor{pastelorange}{HTML}{FFB482}
\definecolor{pastelgreen}{HTML}{8DE5A1}
\definecolor{pastelred}{HTML}{FF9F9B}
\definecolor{pastelpurple}{HTML}{D0BBFF}

\newcommand{\hlred}[1]{\sethlcolor{pastelred}\hl{#1}}
\newcommand{\hlorange}[1]{\sethlcolor{pastelorange}\hl{#1}}
\newcommand{\hlblue}[1]{\sethlcolor{pastelblue}\hl{#1}}
\newcommand{\hlgreen}[1]{\sethlcolor{pastelgreen}\hl{#1}}

\newcommand{\hlredmuted}[1]{\sethlcolor{pastelred!60}\hl{#1}}
\newcommand{\hlorangemuted}[1]{\sethlcolor{pastelorange!60}\hl{#1}}

\newcommand{\Example}[7]{#7\newline $X_{\text{src}}$: #1\newline $X_{\text{tgt}}$: #2\newline $Q$: {\emph{#3}} $A$: #4\newline{\tiny #6} {\tiny #5}}
\def\AnswerableXcGood{\colorbox{pastelgreen!50}{\strut \ding{51} Answerable with $X_{\text{src}}$}}
\def\AnswerableXcMulti{\colorbox{pastelorange!50}{\strut $\sim$ Answerable with $X_{\text{src}}$ (Multi.)}}

\def\AnswerableXsYes{\colorbox{pastelred!50}{\strut \ding{55} Answerable with $X_{\text{tgt}}$}}
\def\AnswerableXsVague{\colorbox{pastelgreen!50}{\strut \ding{51} Vaguely Answerable with $X_{\text{tgt}}$}}
\def\AnswerableXsNo{\colorbox{pastelgreen!50}{\strut \ding{51} Not Answerable with $X_{\text{tgt}}$}}

\def\AccuracyExtCorrect{\colorbox{pastelgreen!50}{\strut \ding{51} $E$ is correct}}
\def\AccuracyExtPartial{\colorbox{pastelorange!50}{\strut $\sim$ $E$ is partial}}
\def\AccuracyExtWrong{\colorbox{pastelred!50}{\strut \ding{55} $E$ is wrong}}

\def\AccuracyAbsCorrect{\colorbox{pastelgreen!50}{\strut \ding{51} $A$ is correct}}
\def\AccuracyAbsPartial{\colorbox{pastelorange!50}{\strut $\sim$ $A$ is partial}}

\def\GivennessGood{\colorbox{pastelgreen!50}{\strut \ding{51} Givenness}}
\def\GivennessBad{\colorbox{pastelred!50}{\strut \ding{55} Givenness}}

\def\LocalizationGood{\colorbox{pastelgreen!50}{\strut \ding{51} $R$ is related}}
\def\LocalizationMissing{\colorbox{pastelorange!50}{\strut $\sim$ $R$ missing}}

\def\LocalizationCorrectlyAbsent{\colorbox{pastelgreen!50}{\strut \ding{51} Deletion = no $R$}}

\def\JargonAAAA{\colorbox{pastelgreen!50}{\strut \ding{51} Jargon (++)}}
\def\JargonAAA{\colorbox{pastelgreen!50}{\strut \ding{51} Jargon (+)}}
\def\JargonAA{\colorbox{pastelred!50}{\strut \ding{55} Jargon (-)}}
\def\JargonA{\colorbox{pastelred!50}{\strut \ding{55} Jargon (-{}-)}}

\def\StandaloneGood{\colorbox{pastelgreen!50}{\strut \ding{51} Standalone}}
\def\StandaloneBad{\colorbox{pastelred!50}{\strut \ding{55} Standalone}}

\def\GPT{\colorbox{gray!20}{\strut \textbf{GPT-4}}}
\def\Llama{\colorbox{gray!20}{\strut \textbf{Llama-2}}}
\def\Mistral{\colorbox{gray!20}{\strut \textbf{Mistral}}}
\def\NLI{\colorbox{gray!20}{\strut \textbf{NLI Pipeline}}}

\title{{\sc InfoLossQA}: Characterizing and Recovering Information Loss in \\Text Simplification}

\author{
Jan Trienes$^{1,2,4}$\thanks{Work done while visiting UT Austin.}\quad
Sebastian Joseph$^3$\quad
Jörg Schlötterer$^{4,5}$\\
{\bf Christin Seifert}$^4$\quad
{\bf Kyle Lo}$^6$\quad
{\bf Wei Xu}$^7$\quad
{\bf Byron C. Wallace}$^8$\quad
{\bf Junyi Jessy Li}$^3$\\
$^1$University of Duisburg-Essen\quad
$^2$Institute for AI in Medicine, University Hospital Essen\\
$^3$The University of Texas at Austin\quad
$^4$University of Marburg\quad
$^5$University of Mannheim\\
$^6$Allen Institute for AI\quad
$^7$Georgia Institute of Technology\quad
$^8$Northeastern University\\
\texttt{jan.trienes@uni-marburg.de}\quad
\texttt{jessy@utexas.edu}}

\begin{document}
\maketitle

\begin{abstract}
Text simplification aims to make technical texts more accessible to laypeople but often results in deletion of information and vagueness.
This work proposes \textsc{InfoLossQA}, a framework to characterize and recover simplification-induced information loss in form of question-and-answer (QA) pairs.
Building on the theory of Questions Under Discussion, the QA pairs are designed to help readers deepen their knowledge of a text.
First, we collect a dataset of 1,000 linguist-curated QA pairs derived from 104 LLM simplifications of English medical study abstracts.
Our analyses of this data reveal that information loss occurs frequently, and that the QA pairs give a high-level overview of what information was lost.
Second, we devise two methods for this task:
end-to-end prompting of open-source and commercial language models, and a natural language inference pipeline.
With a novel evaluation framework considering the correctness of QA pairs and their linguistic suitability, our expert evaluation reveals that
models struggle to reliably identify information loss and applying similar standards as humans at what constitutes information loss.\footnote{Code, dataset and an interactive data viewer is available at \url{https://jantrienes.github.io/ts-info-loss/}.}
\end{abstract}

\section{Introduction}
Technical texts, many of which exist in high-stake domains (e.g., medicine), are often written in a language incomprehensible to lay readers.
Improving the accessibility of such texts may help address wider social issues, e.g., disinformation~\cite{undisinfo} and access to higher education~\cite{Goff:2004:Preferred}.
Automatic text simplification that rewrites text into plain language may therefore be a technology for good.
With the adoption of LLMs, document-level text simplification has significantly progressed in recent years~\cite{August:2023:TOCHI,Laban:2023:ACL,Agrawal:2024:TACL}.

Simplification is an inherently \emph{lossy} process: Even when done by professional editors, the resulting plain language tends to lose details and become more generic~\cite{Li:2015:AAAI}, and some content is omitted~\cite{Zhong:2020:AAAI}.
The \emph{over}-simplification of content---including excessive deletion and vagueness---may lead to reduced comprehension~\cite{Agrawal:2024:TACL} or in its worst case to misinterpretations and factual errors~\cite{Devaraj:2022:ACL}.
Therefore, we consider the following question: \emph{How can we characterize information loss and help readers recover what is lost in an intuitive and understandable manner?}

\begin{figure}[t]
\includegraphics[width=\linewidth]{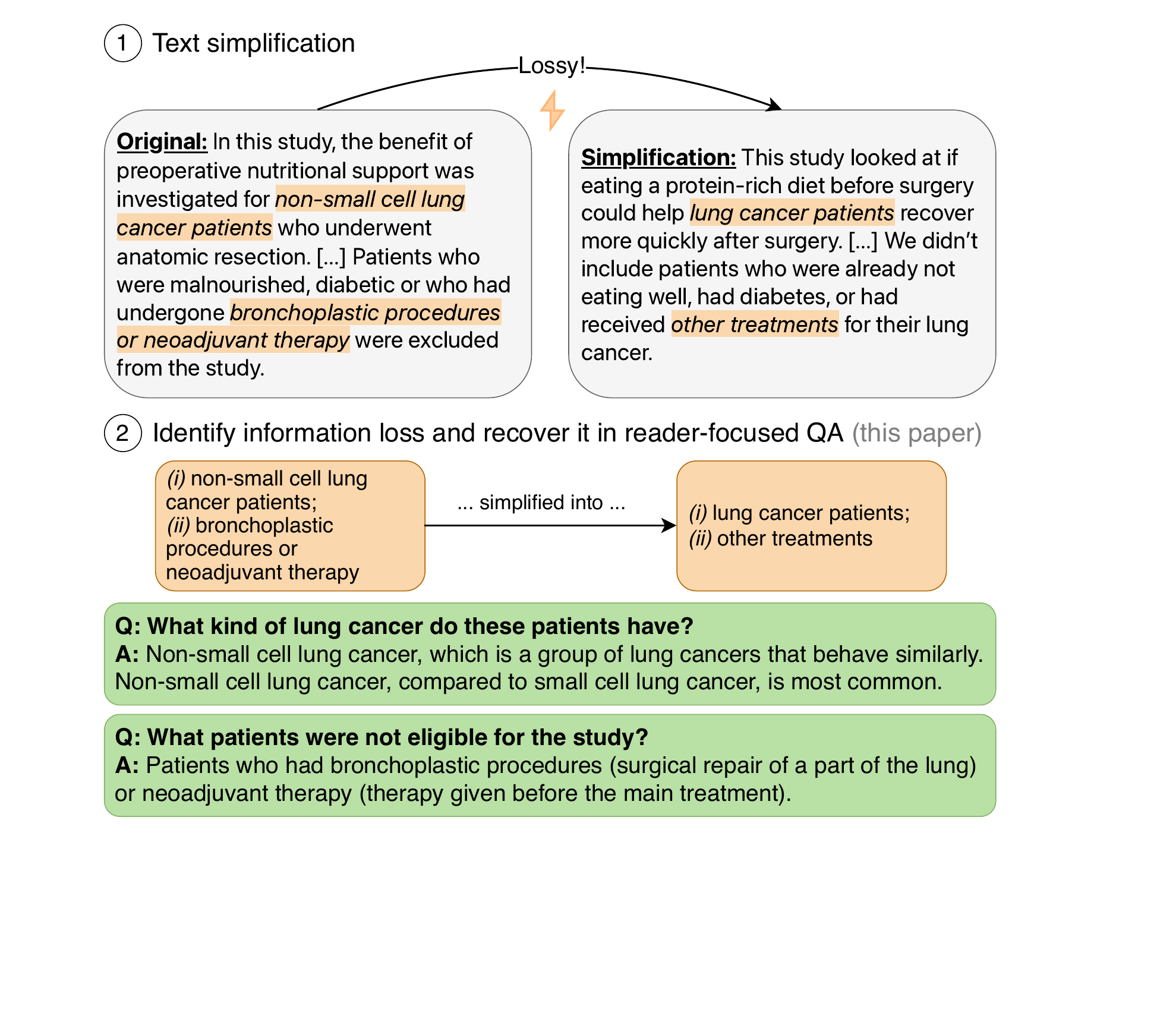}
\caption{The goal of \textsc{InfoLossQA} is to generate a series of QA pairs that reveal to lay readers what information a simplified text lacks compared to its original.}
\label{fig:intro}
\end{figure}

This paper presents a question generation and answering framework to characterize and recover information loss during simplification in a reader-centric way.
Inspired by the theoretical work of ``potential questions''~\cite{Onea:2016:Potential} and expectation-driven Questions Under Discussion (QUD, \citealp{kehler2017evaluating}), information loss is \textbf{characterized} in the form of pragmatically felicitous questions rooted in simplified texts, without assuming access to the original text.
Information loss is then \textbf{recovered} by answering these questions in plain language, based on the original (technical) text (\cref{fig:intro}).
Informed by QUD, the questions are curiosity-driven and information-seeking, and are subject to well-defined linguistic constraints~\cite{Riester:2018:book,Wu:2023:EMNLP}.
While prior work used factoid QAs to describe specific factual changes~\cite{Cole:2023:EACL}, the QUD framework suggests higher-level questions, and allows us to characterize general yet subtle language changes pervasive in factually equivalent texts.
These include lower specificity~\cite{Louis:2011:WS} and greater degrees of vagueness~\cite{VanDeemter:2010:NotExactly}.
We envision the resulting QA pairs being directly incorporated in interactive simplification tools~\cite{Fok:2023:arXiv}.

\paragraph{Contribution 1: Task formulation and dataset.}
We first introduce {\sc InfoLossQA}, a linguist-curated dataset grounded in the scenario of lay readers digesting simplified abstracts of randomized controlled trials (RCTs) in medicine. 
We focus on RCTs because they combine important and illustrative properties: They use highly technical/inaccessible language; lay audiences often having keen interest in their content; and misinterpretations are risky in the healthcare domain.
{\sc InfoLossQA} consists of 1,000 QA pairs across 104 RCT abstracts simplified by GPT-4 (prior work has shown that GPT-3.5 is a capable simplifier of medical texts; \citealp{Shaib:2023:ACL}).
The QA pairs provide a rich description of what information is lost both on a pragmatic and semantic level.
Our analyses show that
questions often elaborate about \emph{why} and \emph{how} an RCT was conducted, going beyond frequently asked questions
in this domain~\cite{August:2023:TOCHI}.

\paragraph{Contribution 2: Automatic methods.}
We then present two methods to perform the question generation and answering task. 
First, an end-to-end model, where we prompt several open-source and API-based LLMs with detailed task instructions and a one-shot example.
Second, a natural language inference (NLI) pipeline which reasons over entailment of atomic facts~\cite{Stacey:2023:arXiv}.

\paragraph{Contribution 3: Evaluation framework and human quality judgments.}
Finally, we design a comprehensive framework to evaluate models at this task.
The framework includes 10 quality desiderata evaluating the correctness of QAs, their linguistic suitability, and the recall of human-written QAs.
Expert judgments are collected on more than 400 generated QA pairs.
We find that models have good question generation and answering capabilities but fall short at reliably identifying information loss.
In this regard, the NLI pipeline is more effective than open-source LLMs, but produces QA pairs with a smaller granularity.

\section{The \textsc{InfoLossQA} Task}
\label{sec:task}
\paragraph{Motivation.} Information loss is inherent to simplification, insofar as the task typically demands producing outputs which are both comparatively uncomplicated \emph{and} reasonably concise.
Characterizing and revealing the elided content is critical to (1) provide a path for users to achieve better understanding than only seeing the simplified text, supporting users' growth~\cite{Gooding:2022:SLPAT}, and (2) to allow researchers to evaluate the quality of a simplified text, given that omitting or oversimplifying key information may yield misleading texts~\cite{Devaraj:2022:ACL}.

\paragraph{Task formulation.}
Given a pair of documents, the original text $X_{\text{src}}$ and the simplified text $X_{\text{tgt}}$, the task entails generating QA pairs that each reveal to lay readers one instance of information loss.
We define an instance of information loss as a tuple $(X_{\text{src}}, X_{\text{tgt}}, Q, A, E, R)$.
A potential question $Q$ asks for information absent from $X_{\text{tgt}}$.
The answer $A$ 
provides the missing information.
We additionally require two forms of localization or grounding to support richer analysis and scrutiny of errors in the generated QA pairs.
First, each answer $A$ must be supported by \emph{evidence} $E \in X_{\text{src}}$ extracted from the original text.
Second, each QA pair may be supported by an extractive \emph{rationale} $R \in X_{\text{tgt}}$ which localizes where the information loss or vagueness occurs within the simplified text.
Note that while $E$ always exists, 
$R$ may not (some information in $X_{\text{src}}$ may be completely absent from $X_{\text{tgt}}$).

This formulation permits multiple information loss instances for a given $X_{\text{src}}$ and $X_{\text{tgt}}$ pair.
We do not constrain the length or linguistic unit(s) of the extractive spans ($E$ and $R$): they can comprise words, phrases, sentences, or entire paragraphs, and a single $E$ or $R$ can be one or more spans.

\paragraph{Types of information loss.}
For the purposes of this study, we define two types of information loss:
\begin{enumerate}[noitemsep,topsep=0pt]
\item \textbf{\hlred{Deletion.}} Pieces of information which were not included in the simplification.
\item \textbf{\hlorange{Oversimplification.}} Pieces of information that were simplified to the extent that they are vague or devoid of their original meaning. This is where the rationale $R$ is identified.
\end{enumerate}
These categories are meant to be flexible to capture a variety of information transformations, but also concrete to be consistently applied in an annotation protocol (\cref{sec:dataset}).

\paragraph{Linguistic suitability.}
One of our primary goals is to generate QA pairs that may enhance users' comprehension of a text going beyond its simplified version~\cite{Fok:2023:arXiv}.
To this end, we pose two requirements for the QA pairs:

\textit{(1) Readability.}
The language level of the questions and answers should match that of the simplified text.
That is, while the QA must discuss technical material from the original text, it should be explained at a level appropriate for the reader.\footnote{It is possible that $X_{\text{tgt}}$ itself is not at the right level for a given reader. However, this is out of scope of our study.}

\textit{(2) Givenness.}
Questions should be pragmatically felicitous to be understood by a reader without having seen the answer or the original text.
The theory of QUD formalizes this through the Givenness constraint~\cite{Riester:2018:book,Wu:2023:EMNLP} which specifies that $Q$ should not contain concepts that are hearer-new~\cite{Markert:2012:ACL} with respect to a common ground. Here, the common ground is the simplified text $X_{\text{tgt}}$.
Intuitively, this means that it should be clear from reading the question how the answer would expand on what a reader already knows from the simplified text.

\cref{fig:task-example} illustrates the different elements and challenging nature of the task.
To identify information loss, models cannot rely on lexical overlap, ordering, or other surface-level properties.

\begin{figure}[t]
\small
\fbox{\parbox{.98\linewidth}{
\setlength{\parskip}{0.1\baselineskip}
\textbf{Original ($X_{\text{src}}$):} These results indicate that \hlred{acute/chronic endurance ($E_1$)} arm-cranking with EMS applied to the lower limbs \hlorange{improves the brachial artery endothelial function ($E_2$)} more markedly than the same exercise without EMS.
\\[-0.8\baselineskip]

\textbf{Simplified ($X_{\text{tgt}}$):} The study concluded that doing the arm-cranking exercise with EMS on the lower body can \hlorange{improve arm function ($R_2$)} more than without it. 
\\[-0.8\baselineskip]

$Q_1$: On what timeframes does EMS improve training? $A_1$: It helps in single (acute) or repeated applications (chronic).
\\[-0.8\baselineskip]

$Q_2$: How did the researchers measure how well EMS works?
$A_2$: They measure to what extent the main artery of the arm widens, which is called flow-mediated vasodilation.
\\[-0.8\baselineskip]

$Q_1'$ (\Lightning): Did EMS improve training in acute and chronic applications?}}
\vspace{-0.5em}
\caption{Example with
a \hlred{Deletion} (``acute/chronic'') and an \hlorange{Oversimplification} (``improve arm function'' is too broad given that EMS improves ``artery function'').
These give rise to two QA pairs ($Q_1$ and $Q_2$) which fulfill the Readability and Givenness constraints. For contrast,
$Q_1'$ violates (\Lightning) givenness. $Q_1$ is likely more natural to lay readers because it could be asked without having seen the original text (no presupposition that the study looked at short-term and long-term effects).}
\label{fig:task-example}
\vspace{-0.5em}
\end{figure}

\section{Data Collection}
\label{sec:dataset}
\begin{figure*}[t]
\centering

\begin{subfigure}{0.6\textwidth}
\centering
\includegraphics[width=1\textwidth]{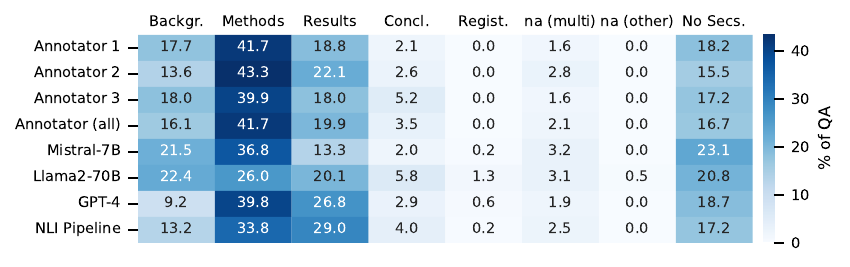}
\caption{Distribution of QA pairs over RCT abstract sections. Some QAs are localized in multiple sections (multi.), in a rare section (other) or their abstracts do not have a section structure (19.2\% of abstracts; no secs.).}
\label{fig:info-loss-distribution-section}
\end{subfigure}
\hfill
\begin{subfigure}{0.38\textwidth}
\centering
\includegraphics[width=1\linewidth]{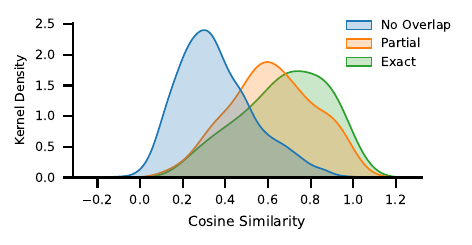}
\caption{Cosine similarity of annotators' questions ($Q$) by whether their evidence spans ($E$) have no/partial/complete overlap.
}
\label{fig:question-similarity}
\end{subfigure}

\begin{subfigure}{\textwidth}
\includegraphics[width=\textwidth]{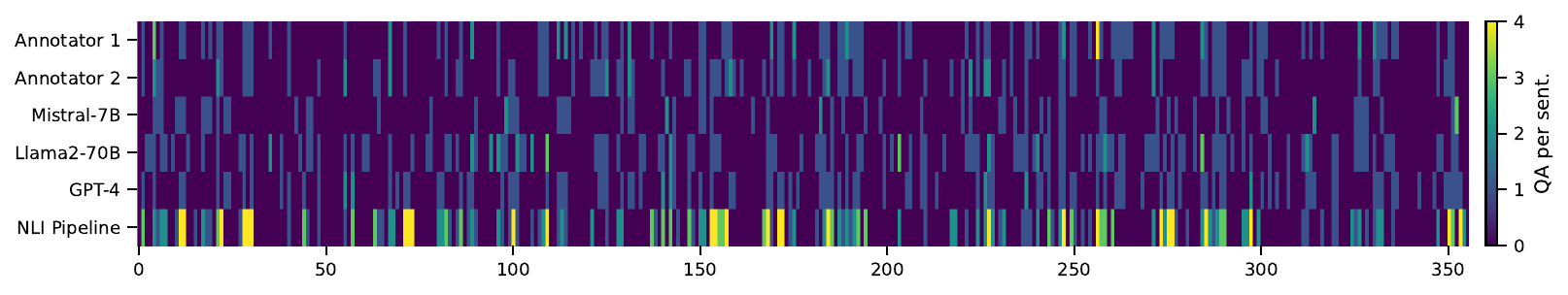}
\caption{Number of QA pairs per sentence for a random selection of 25 documents. The x-axis are sentence indices.
}
\label{fig:info-loss-distribution-sentence}
\vspace{-0.5em}
\end{subfigure}
\caption{Distribution of information loss. Humans produce a similar distribution of questions by section \emph{(a)}, but the questions differ in their localization \emph{(c)}. A similar localization results in more similar questions \emph{(b)}. Comparing humans to models, we see differences where questions are localized, and by extension also what they are about.}
\label{fig:info-loss-distribution}
\vspace{-0.5em}
\end{figure*}

\subsection{Simplification Corpus}
We focus on medical texts, which provide an important and illustrative case where lay individuals may be keenly interested in understanding newly published science.
Specifically, we consider abstracts of medical journal articles that describe the conduct and findings of RCTs. RCTs are the foundation of evidence-based medicine that informs patient care~\cite{Sackett:1998:Evidence}.
But those articles use technical language, which is effectively inaccessible to lay patients~\cite{Zuccala:2010:Open,August:2023:TOCHI}.
This means that individuals may already interact with LLMs to simplify such texts.
But automated simplification of RCTs may result in information loss, which risks readers misinterpreting findings.

To better quantify information loss in LLM-simplifications, we create a parallel simplification corpus with GPT-4 (\texttt{\small gpt-4-0613}).\footnote{No curated parallel corpus exists for RCT simplification.}
On the related task of plain language summarization of full RCTs, \citet{Shaib:2023:ACL} observed low error rates (<10\% of summaries affected), but we do not yet have a quantitative understanding of information loss in LLM-simplifications.
We sample RCT abstracts from the {\tt\small Evidence Inference v2.0} dataset~\cite{DeYoung:2020:BioNLP,Lehman:2019:NAACL}, originally sourced from PubMed.
\cref{tab:prompts-overview} provides the simplification prompt and example outputs.

\subsection{Annotation Process}\label{sec:annotationprocess}
\label{sec:annotation-process}
The process for annotating \textsc{InfoLossQA} is as follows.
First,  annotators examine both the original and simplified versions of the text.
Next they identify and highlight information loss spans.
For each highlighted instance, annotators compose a QA pair.
We used the {\tt\small Thresh} toolkit for annotation~\cite{Heineman:2023:EMNLP}.
See \cref{sec:appendix-annotation-guidelines} for annotation guidelines and interface screenshots.

\paragraph{Scenario grounding.}
Technical texts like RCTs have a long tail of information that can be lost in simplification (e.g., numerical results and significance values).
Annotating \emph{all} information loss is intractable, and the resulting QAs would likely overwhelm readers.
Therefore, we grounded annotation in a scenario. Annotators were asked to imagine a patient who could benefit from better understanding the RCT at hand, and to focus on information that may be salient in this scenario.

This notion of saliency is subjective as people have different information needs.
Similarly, prior work in QUD~\cite{Ko:2022:EMNLP} showed that question generation by human annotators is also subjective: People ask different questions even when the answer to these questions is the same.
In our dataset, each document was annotated by two annotators, independently. This allows us to evaluate the extent to which people ask similar questions (\cref{sec:analysis}).

\paragraph{Annotators.}
We hired three expert annotators who are native English speakers, majoring in linguistics and experienced in annotating medical texts.
To ensure high annotation quality, we conducted a pilot phase with written feedback and held frequent review meetings to discuss questions.
Annotators were compensated \$15/hr.\footnote{Higher than local minimum wage at the time of study.}
The median duration per document was 22 minutes.

\paragraph{Dataset statistics.} We collected annotations on 104 abstract-simplification pairs (average number of  tokens: $|X_{\text{src}}| = 312$; $|X_{\text{tgt}}| = 271$) yielding 1,000 questions and answers (average number of tokens: $|Q| = 11.4$; $|A| = 30.3$).\footnote{Tokenizing with \textsc{nltk}~\cite{Bird:2009:NLTK}.}
Each annotator wrote on average between 4.1 and 5.7 QA pairs (i.e., information loss instances) per document.
We observe that oversimplification is more prevalent (74.1\% of questions) in this corpus than complete deletion of information (25.9\% of questions).
Additional statistics are given in \cref{tab:stats-breakdown}.

\subsection{Analysis of \textsc{InfoLossQA}}\label{sec:analysis}
We analyze the fine-grained annotation of questions and their localization in form of evidence spans
to get a better understanding of how models preserve (or omit) content during simplification.

\begin{table*}[t]
\small
\rowcolors{2}{gray!10}{white}
\renewcommand{\arraystretch}{1.1}
\resizebox{\textwidth}{!}{%
\begin{tabular}{p{0.32\textwidth}rrrp{0.5\textwidth}}
\toprule
\textbf{Question Type} & \textbf{\% All} & \textbf{\% O} & \textbf{\% D} & \textbf{Examples} \\
\midrule
\textbf{Procedural.} Asking for the procedures, tools, or methods by which a certain outcome is achieved. & 34.3 & 34.1 & 34.7 &
\textbullet\ How did they measure the patients' opioid medication needs?\newline
\textbullet\ How did the study control for bias?
\\
\textbf{Concept.} Asking for a definition of an event or a concept. & 25.7 & 28.1 & 18.9 &
\textbullet\ What kind of hip surgery were patients undergoing?
\textbullet\ What type of mental illnesses are being studied?
\\
\textbf{Extent.} Asking for the extent or quantity of an event or a concept. & 17.0 & 19.2 & 10.8 &
\textbullet\ How many patients were in each group?
\textbullet\ How much lurasidone was given to the patients in the trial?
\\
\textbf{Comparison.} Asking for comparison among multiple events or concepts. & 8.3 & 8.9 & 6.6 &
\textbullet\ How much did the abnormal blood vessels reduce in group 1 compared to group 2?
\textbullet\ How did pre-meal insulin affect platelet activation compared to the placebo?
\\
\textbf{Consequence.} Asking for the consequences or results of an event. & 5.7 & 3.0 & 13.5 &
\textbullet\ What else does increased blood flow do?
\textbullet\ What was the correlation between measures for total daily calcium intake?
\\
\textbf{Cause.} Asking for the cause or reason for an event or a concept. & 4.3 & 1.6 & 12.0 &
\textbullet\ What was a motivation behind the study?
\textbullet\ Why are soy products being looked at in this study?
\\
\textbf{Example.} Asking for example(s) or instance(s) of an event or a concept. & 3.4 & 4.3 & 0.8 &
\textbullet\ What are the other brain functions that were tested?
\textbullet\ What medication is typically used for outpatients?
\\
\textbf{Disjunction.} Asking for the true one given multiple events or concepts, where comparison among options is not needed. & 0.6 & 0.5 & 0.8 &
\textbullet\ Which one of the three components did this group's supplement contain?
\textbullet\ Which gel saw the increase in beneficial microbes after 24 hours?
\\
\textbf{Verification.} Asking for the truthfulness of an event or a concept. & 0.7 & 0.3 & 1.9 &
\textbullet\ Were there any other changes in blood sugar level observed in the study?
\textbullet\ Were both eyes of each patient tested?
\\
\textbf{Judgmental.} Asking for the opinions of the answerer's own. & \multicolumn{3}{c}{\emph{Not observed.}} & \\
\bottomrule
\end{tabular}}
\caption{Example questions in \textsc{InfoLossQA} classified according to the question taxonomy by~\citet{Cao:2021:ACL}. Percent of all questions is given by category (O = Oversimplification; D = Deletion).}
\label{tab:question-taxonomy}
\vspace{-0.5em}
\end{table*}

\begin{table}[t]
\small
\centering
\begin{tabular}{lrrr}
\toprule
\textbf{Granularity (Measure)} & \textbf{Del.} & \textbf{Over.} & \textbf{All} \\
\midrule
Span, exact (F1) & 0.20 & 0.13 & 0.17 \\
Span, partial (F1) & 0.34 & 0.34 & 0.34 \\
Sentence (Krippendorff's $\alpha$) & -- & -- & 0.50 \\
\bottomrule
\end{tabular}
\vspace{-0.5em}
\caption{Evidence span ($E$) overlap between annotators.
Del = Deletions; Over = Oversimplification.
}
\label{tab:iaa-annotation}
\vspace{-0.5em}
\end{table}

\paragraph{Where in the text is most information lost?}
We combine the precise localization of a QA pair with structural elements of the abstracts to identify areas where the most information loss occurs.
By convention, most RCT abstracts are structured, i.e., having section headings for Background, Methods, Results and Conclusion (most commonly).
We use a regular expression to identify the start and end boundaries of sections.
Because the specific naming of such sections differs across articles, we collected a lookup dictionary to normalize synonymous headers.\footnote{For example, ``Design,'' ``Methods'' and ``Materials and Methods'' are all mapped to ``Methods.'' }
Afterwards, we align QA pairs to sections based on their localization.

\cref{fig:info-loss-distribution} shows the distribution of QAs over RCT sections.
Most information loss occurs in the Methods section (41.7\% of QAs).
This is intuitive as Methods sections are likely to be particularly technical.
The least information loss occurs in the Conclusion (3.5\% of QAs) section.
A small percentage of QAs spans multiple sections. These tend to be about concepts that were repeatedly oversimplified.

\paragraph{What types of questions are asked?}
To characterize the information that is typically lost, we categorize questions according to the 10 pragmatic function types defined by \citet{Cao:2021:ACL} using a few-shot prompting classifier.\footnote{Prompting \texttt{gpt-4-0613} with annotation guidelines of \citet{Cao:2021:ACL} and few-shot examples achieves an F1 score of 0.87 on a validation set (\cref{tab:prompts-overview}).}
\cref{tab:question-taxonomy} shows the question distribution across types, their definition and examples.
Most questions are of type \emph{procedural}, requesting specific details about how the study was conducted (34.3\%).
This aligns with the earlier observation that most information loss occurs in Methods sections.
The second most common type of questions seek more concrete definitions for overgeneralized \emph{concepts} (25.7\%), such as the health condition or clinical procedure.
Comparing deletions and oversimplifications, we observe a similar distribution, with the exception of questions asking for the \emph{consequences} (e.g., experimental results) or \emph{causes} (e.g., what motivated the study) of an event being more prevalent among deletions.

\cref{sec:appendix-faqs} also provides a domain-specific classification of QAs in terms of frequently asked questions about RCTs~\cite{August:2023:TOCHI}.

\paragraph{How similar are information needs across annotators?}
As discussed in \cref{sec:annotationprocess}, the {\sc InfoLossQA} task is subjective.
Enlisting two annotators per document allows us to quantitatively study information needs of different individuals.

First, we analyze to what extent annotators highlight the same evidence spans $E$. 
We calculate three measures: (1) \textbf{strict span-level F1}, where boundaries and type (deletion vs. oversimplification) have to match exactly,
(2) \textbf{partial span-level F1} where two spans are considered a match if their token-level intersection over union exceeds a threshold~\cite{DeYoung:2020:EMNLP},\footnote{We tokenize on whitespace and use a threshold of $t=0.5$.} and (3) \textbf{sentence-level Krippendorff's $\alpha$} where we project span-level annotations to a binary label indicating if a sentence has at least one span~\cite{Goyal:2022:EMNLP}.
The sentence-level $\alpha$ shows that the annotators do tend to pick up information loss localized in the same evidence \emph{sentence} to a moderate degree~\cite{Artstein:2008:CL}.
However, the precise evidence span (in smaller linguistic units) is more subjective, as indicated by the lower span-level scores (\cref{tab:iaa-annotation}).

Second, we analyze the similarity of questions as a function of whether their underlying evidence spans $E$ have complete, partial, or no overlap.
We use cosine similarity over question embeddings~\cite{Reimers:2019:EMNLP}.
Generally, questions become more similar when the evidence overlaps (\cref{fig:question-similarity}).
However, subjectivity remains: Some questions are different although the evidence is the same. Similar observations were made for QUD generation~\cite{Ko:2022:EMNLP}.

\section{Automatic Information Loss Detection}
No system in literature is directly applicable to solve all parts of \textsc{InfoLossQA}.
Therefore, we propose and evaluate two systems: (1) End-to-end prompting, contrasting several strong LLMs; (2) An NLI pipeline.
We describe the methods below and provide detailed parameters in \cref{tab:prompts-overview}.

\paragraph{Method 1: LLM prompting.}
We design a one-shot prompt that specifies the task, closely resembling the manual annotation guidelines.
We ask all models to localize information loss instances, to provide a short motivation for why it is considered information loss, akin to Chain-of-Thought prompting~\cite{Wei:2022:NIPS}, and finally to generate the QA pairs.
We benchmark three LLMs:
\textbf{Mistral-7B}, \textbf{Llama2-70B}, and \textbf{GPT-4}.\footnote{Specifically, we use \texttt{gpt-4-0613}, \texttt{llama-2-70b-chat}, and \texttt{mistral-7b-instruct-v0.1}.}
All prompts were manually tuned for each model, though this was a heuristic process and better prompts may exist.

\paragraph{Method 2: NLI pipeline.}
To contrast with end-to-end LLMs, we implement a three-step pipeline: (i) fact extraction, (ii) entailment classification, (iii) QA generation for facts with a \emph{neutral} relation.
NLI assesses if a ``hypothesis'' is inferable from a ``premise,'' categorizing it as \emph{entailed}, \emph{contradictory}, or \emph{neutral}.
We treat the original text $X_{\text{src}}$ as hypothesis, and the simplification $X_{\text{tgt}}$ as premise.
Intuitively, a neutral relation indicates information loss, where at least one piece of information in $X_{\text{src}}$ cannot be inferred from $X_{\text{tgt}}$.
In contrast, entailment indicates faithful conveyance of all information.

Information loss can be as granular as a word or phrase, and one sentence may contain multiple distinct instances.
Therefore, we adopt NLI over fine-grained facts~\cite{Stacey:2023:arXiv}.
Formally, we decompose each sentence in $X_{\text{src}}$ into atomic facts using an LLM~\cite{Stacey:2023:arXiv}.
Then, a standard NLI model~\cite{Laurer:2024} classifies entailment of each fact against $X_{\text{tgt}}$.
Finally, we prompt an LLM to generate QA pairs given $X_{\text{src}}$, $X_{\text{tgt}}$, and the list of neutral facts.
The sentence that generated the fact serves as evidence $E$.\footnote{We do not implement localization $R$ in the simple text, but note that interpretable NLI models could provide such a localization~\cite{Stacey:2022:EMNLP,Rodriguez:2023:arXiv}.}

\section{Evaluation Framework}
\label{sec:evaluation}
We next propose a comprehensive framework to evaluate automatic systems for \textsc{InfoLossQA}.
All evaluation is done manually by trained annotators.
We describe the evaluation criteria below and \cref{sec:appendix-evaluation-guidelines} provides full annotator guidelines.

\paragraph{Aspect 1: Correctness of generated QA pairs.}
Our first criterion, \textbf{Answerability} (3-point), states that the generated \emph{questions} ($Q$s) should be about an information loss.
Valid questions must be answerable with the original text (i.e., there exists an evidence $E$ answering $Q$), but unanswerable or only vaguely answerable with the simple text.

The second criterion, \textbf{Answer Accuracy}, states that questions should be correctly answered by the extracted and generated answers.
We evaluate this in three parts. First, \textbf{Accuracy - Evidence} (3-point) captures how well the highlighted evidence $E$ answers $Q$.
Second, \textbf{Accuracy - Answer} (3-point) evaluates how well $A$ answers $Q$.
Finally, we ask if $A$ contains any \textbf{Hallucinations} (binary).

\paragraph{Aspect 2: Linguistic suitability.}
We aim for QAs that are \emph{pragmatically felicitous} and \emph{comprehensible} for readers (cf. \cref{sec:task}).
We propose evaluating this through four linguistic criteria.
First, we evaluate \textbf{Givenness} (binary), closely following the constraints of QUD~\cite{Riester:2018:book,Wu:2023:EMNLP}: A question should only contain concepts that are directly mentioned in the question context, inferable, or generally known.
We define the question context as the entire simple text $X_{\text{tgt}}$ for QAs on Deletions, and everything up to and including the rationale span $R$ for QAs on Oversimplifications.\footnote{%
With this definition we intend to enable interactions where users can inspect oversimplification QAs \emph{as they read} the text, e.g., visualized as expandable highlights~\cite{Fok:2023:arXiv}.
}
Second, when a question seeks clarification about an oversimplified concept, there should be a rationale $R$ in $X_{\text{tgt}}$ as an opportunity to show users that the question addresses said vagueness (\textbf{Rationale}, 3-point).
Third, we evaluate if $A$ contains any unexplained \textbf{Jargon} (4-point).
Finally, \textbf{Standalone} (binary) states that answers must be decontextualized~\cite{Choi:2021:TACL}, i.e., they should not contain unresolved coreferences or acronyms that would require reading the original text.

\paragraph{Aspect 3: Recall of human QA.}
Lastly, we evaluate if models identify similar information loss as the references in \textsc{InfoLossQA}.
A challenge in this evaluation is that both humans and models produce a \emph{set of QAs} without a strict 1-to-1 correspondence.
We found that directly comparing two question sets is too cognitively demanding with typical sets of more than five QAs.
Therefore, we opt for a pairwise setting, comparing each reference QA with one model QA at a time, and marking the reference as either \emph{fully recalled}, \emph{partially recalled} or \emph{not recalled}.
As comparing all possible pairs is intractable, we exclude unlikely matches following a paraphrase mining approach~\cite{Wright:2022:EMNLP}.
We report the detailed procedure to estimate a minimum similarity threshold in \cref{sec:appendix-evaluation}.

\paragraph{Evaluation process.}
We sample 100 QAs per model for aspect 1+2, and 50\% of the documents for aspect 3.
All samples were independently evaluated by two of the experts described in \cref{sec:annotation-process}.
Evaluation took 64 hours for aspect 1+2, and 20 hours for aspect 3, with additional training hours.
We measure Randolph's Kappa~\cite{Randolph:2005} for inter-evaluator agreement and observe a substantial average agreement of $\kappa = 0.70$ (\cref{tab:iaa-evaluation}).
\cref{sec:appendix-evaluation} provides details on sampling and exclusion of invalid LLM generations.

\begin{table*}[t]
\small
\setlength{\tabcolsep}{3pt}
\centering
\begin{tabularx}{\textwidth}{
    l
    >{\centering}X
    >{\centering}X
    >{\centering}X
    >{\centering}X
    >{\centering}X
    >{\centering}X
    >{\centering}X
    >{\centering}X
    >{\centering}X
    >{\centering}X
    >{\centering}X
    >{\centering\arraybackslash}X
}
\toprule
    & \multicolumn{3}{c}{\textbf{$Q$ is Answerable w/ $X_{\text{src}}$}} & \multicolumn{3}{c}{\textbf{$Q$ is Answerable w/ $X_{\text{tgt}}$}} & \multicolumn{3}{c}{\textbf{Accuracy - Evidence ($E$)}} & \multicolumn{3}{c}{\textbf{Accuracy - Answer ($A$)}} \\
    \cmidrule(lr){2-4}
    \cmidrule(lr){5-7}
    \cmidrule(lr){8-10}
    \cmidrule(lr){11-13}
\textbf{Model} & Yes & Multi.$^\downarrow$ & No$^\downarrow$ & No & Vague & Yes$^\downarrow$ & Correct & Part.$^\downarrow$ & Wrong$^\downarrow$ & Correct & Part.$^\downarrow$ & Wrong$^\downarrow$ \\
\midrule
Mistral-7B & 72.5 & 19.6 & 7.8 & 32.4 & 30.4 & 37.3 & 74.5 & 12.7 & 12.7 & 84.8 & 12.3 & 2.9 \\
Llama2-70B & 83.0 & 9.7 & 7.3 & 21.4 & 32.5 & 46.1 & 77.7 & 11.7 & 10.7 & 85.9 & 10.7 & 3.4 \\
GPT-4 & 87.5 & 12.5 & \bfseries 0.0 & \bfseries 58.5 & \bfseries 33.0 & \bfseries 8.5 & \bfseries 98.0 & \bfseries 2.0 & \bfseries 0.0 & \bfseries 94.0 & \bfseries 6.0 & \bfseries 0.0 \\
NLI Pipeline & \bfseries 89.7 & \bfseries 7.8 & 2.5 & 49.5 & 25.0 & 25.5 & 77.5 & 4.4 & 18.1 & 77.9 & 20.6 & 1.5 \\
\bottomrule
\end{tabularx}
\vspace{-0.7em}
\caption{Correctness of generated QA pairs as percent of ratings given by two independent annotators over 100 QA pairs per model. Multi. = Multiple valid answers; Part. = Partially complete answer; $^\downarrow$ = lower is better.}
\label{tab:results-part1}
\vspace{-0.7em}
\end{table*}

\begin{table*}[t]
\small
\setlength{\tabcolsep}{3pt}
\centering
\resizebox{\textwidth}{!}{%
\begin{tabular}{lccccccccccccc}
\toprule
    & \textbf{Given. ($Q$)} & \multicolumn{3}{c}{\textbf{Rationale ($R$)}} & \multicolumn{4}{c}{\textbf{Jargon ($A$)}} & \textbf{Stand. ($A$)} & \textbf{Hallu. ($A$)} & \multicolumn{3}{c}{\textbf{Recall}} \\
    \cmidrule(lr){2-2}
    \cmidrule(lr){3-5}
    \cmidrule(lr){6-9}
    \cmidrule(lr){10-10}
    \cmidrule(lr){11-11}
    \cmidrule(lr){12-14}
\textbf{Model}  & \% Good & Good & Unrel.$^\downarrow$ & Miss.$^\downarrow$ & (++) & (+) & (-)$^\downarrow$ & (-{}-)$^\downarrow$ & \% Good & \% Good & Full & Partial & No \\
\midrule
Mistral-7B & 74.5 & \bfseries 52.0 & 5.9 & \bfseries 42.2 & 21.6 & 9.3 & 13.2 & 55.9 & 84.3 & 96.1 & 14.2 & 4.1 & 81.8 \\
Llama2-70B & 75.2 & 34.5 & 5.3 & 60.2 & 26.2 & 7.8 & 17.5 & 48.5 & 79.6 & 92.2 & 16.2 & 4.4 & 79.3 \\
GPT-4 & \bfseries 82.0 & \bfseries 52.0 & 1.5 & 46.5 & 15.0 & 24.0 & \bfseries 11.0 & 50.0 & \bfseries 90.5 & \bfseries 99.0 & \bfseries 28.1 & 3.4 & 68.4 \\
NLI Pipeline & 58.3 & 17.6 & \bfseries 0.0 & 82.4 & \bfseries 28.9 & \bfseries 29.9 & 12.7 & \bfseries 28.4 & 86.3 & 98.0 & 22.9 & 20.0 & \bfseries 57.2 \\
\bottomrule
\end{tabular}
}
\vspace{-0.7em}
\caption{Linguistic suitability of generated QA pairs and recall of human QAs. Given. = Givenness; Stand. = Standalone; Hallu. = Hallucinations; Unrel. = Unrelated; Miss = Missing.
$^\downarrow$ = lower is better.
}
\label{tab:results-part2-3}
\vspace{-0.8em}
\end{table*}

\section{Results and Discussion}
Overall we find that \textsc{InfoLossQA} is a challenging task.
While all models have good question-generation/answering abilities, the main difficulty lies in reliably identifying information loss and ensuring the answers are comprehensible for readers.
Furthermore, models identify different information loss than humans.
Quantitative results are given in \cref{tab:results-part1,tab:results-part2-3} and qualitative examples in \cref{fig:examples}.

\paragraph{Models generate valid questions but not all are about an information loss (\cref{tab:results-part1}).}
The majority of questions is answerable with $X_{\text{src}}$, showing that all models are good at document-grounded question generation.
However, a sizable number of questions lack specificity as they have multiple valid answers (7.8\%--19.6\%).
For example, \emph{``What are the details of the experiments?''} (Ex. 1 in \cref{fig:examples}).
Furthermore, a small number of questions is unanswerable with $X_{\text{src}}$.
Qualitatively, these questions tend to ask for lay explanations of technical terms which is out of scope of information loss.

While most questions are valid, all models could improve at generating questions that are \emph{not} or only \emph{vaguely} answerable with $X_{\text{tgt}}$.
See example 2 (\cref{fig:examples}), where the generated QA does not add any new information.
Notably, the fact-based NLI pipeline outperforms open-source LLMs in identifying unanswerable questions, indicating a promising direction for addressing information loss.

Considering answer accuracy, GPT-4 provides near perfect results both in extracting the correct evidence span (98\% correct) and in rewriting it into a full answer (94\% correct).
Surprisingly, Mistral and Llama2 more often generate correct abstractive answers than extracting the right evidence, implying internally inconsistent LLM responses.
Lastly, across all models, we only observe a small rate of hallucinations (worst: Llama2, 7.8\%).

\paragraph{Models often fail to localize QAs in the simple text (\cref{tab:results-part2-3}).}
Too often models do not produce a rationale span in $X_{\text{tgt}}$, which shows a general limitation of models to correctly discern between deletion and oversimplification (42.2\%--82.4\% missing rationale).
For reference: 25\% of human QAs are deletions, whereas the models classify more than 50\% of QAs as deletions (\cref{tab:stats-breakdown}).
Annotators have substantial agreement for when there should be a rationale span ($\kappa = 0.80$, \cref{tab:iaa-evaluation}), so this could be alleviated by model fine-tuning.

\paragraph{QAs are well-framed, with room for improvement in terms of jargon (\cref{tab:results-part2-3}).}
Considering the Givenness constraint, the end-to-end models provide well-framed questions that only contain concepts familiar to readers and do not have any answer leakage.
This result is on par or slightly better than recent QUD parsers~\cite{Wu:2023:EMNLP}.

Concerning the answers, most can be understood without referring to the original text, i.e., they are Standalone.
Qualitatively, undefined abbreviations are frequent reasons for answers to not be standalone (Ex. 3 in \cref{fig:examples}) which could be solved with a dedicated decontextualization step~\cite{Choi:2021:TACL}.
Lastly, many answers contains unexplained Jargon (-{}-, best NLI 28.4\%, worst Mistral 55.9\%).
Despite instructing all models to generate simple answers, the task likely has too many constraints for a single end-to-end prompt.
By decoupling information loss detection and QA generation, the NLI pipeline can better follow this constraint.

\paragraph{Models have a low recall of human QAs (\cref{tab:results-part2-3}).}
The NLI pipeline is most effective in this criterion (42.9\% of QAs at least partially recalled), but also generates substantially more QAs per document than humans (16.5 vs. 9.6, \cref{tab:stats-breakdown}).
Qualitatively, the generated questions each cover a smaller part of missing content, whereas humans aggregate related missing information into one larger question (see \cref{fig:compositionality-example} for an example).
This is corroborated by the NLI pipeline having the highest \emph{partial} recall of human QAs.
Exciting future directions will be to (1) get models to apply similar standards as humans at what constitutes an information loss, and (2), to study how to aggregate smaller-scope questions into broader questions.

\paragraph{Automatic evaluation: prompting LLMs to evaluate linguistic suitability is difficult (\cref{tab:iaa-evaluation}).}
The previous results rely on manual evaluation by trained annotators, which is expensive and not scalable.
Therefore, we run a preliminary investigation of automatic LLM-based evaluation.
For each of the 10 evaluation criteria, we created a prompt with evaluation instructions, the generated QA pair, and additional context for assessment.
The model outputs a brief rationale followed by the rating, akin to chain-of-thought~\cite{Wei:2022:NIPS}.
No further prompt engineering was done.
Agreement between human ratings and ratings by representative LLMs (GPT-4o and Llama3-70B)\footnote{Specifically, we use the checkpoint \texttt{gpt-4o-2024-05-13}\\ and \texttt{meta-llama/Meta-Llama-3-70B-Instruct}. Full evaluation suite is provided in code release.} are given in~\cref{tab:iaa-evaluation}.

On average, Llama3-70B and GPT-4o agree with humans evaluators to a \emph{fair} and \emph{moderate} degree~\cite{Artstein:2008:CL}.
GPT-4o matches human agreement on criteria related to the correctness of QAs and recall of human QA, suggesting it could substitute human evaluation.
In contrast, LLM-based evaluation of the linguistic suitability of generated QAs appears unreliable and likely requires a more sophisticated method to achieve good agreement with human rating, corroborating results on general QUD evaluation~\cite{Wu:2023:EMNLP}.

\begin{table}[t]
\centering
\small
\setlength{\tabcolsep}{3pt}
\begin{tabular}{lrrr}
\toprule
\textbf{Criterion} & \textbf{Human} & \textbf{GPT-4o} & \textbf{Llama3} \\
\midrule
\multicolumn{3}{l}{\emph{Aspect 1: Correctness of QA}} \\
$Q$ is Answerable w/ $X_{\text{src}}$ & 0.61 & 0.68 & 0.66 \\
$Q$ is Answerable w/ $X_{\text{tgt}}$ & 0.50 & 0.33 & 0.24 \\
Accuracy - Evidence ($E$) & 0.71 & 0.65 & 0.52 \\
Accuracy - Answer ($A$) & 0.68 & 0.63 & 0.57 \\
Hallucinations ($A$) & 0.89 & 0.81 & 0.72 \\
\addlinespace
\multicolumn{3}{l}{\emph{Aspect 2: Linguistic Suitability}} \\
Givenness ($Q$) & 0.79 & 0.30 & 0.24 \\
Rationale Localization ($R$) & 0.80 & 0.18 & 0.06 \\
Jargon ($A$) & 0.57 & 0.34 & 0.43 \\
Standalone ($A$) & 0.69 & 0.21 & -0.30 \\
\addlinespace
\multicolumn{3}{l}{\emph{Aspect 3: Recall of Human QA}} \\
Recall & 0.70 & 0.67 & 0.47 \\
\midrule
Average & 0.70 & 0.48 & 0.36 \\
\bottomrule
\end{tabular}
\caption{
    Inter-evaluator agreement as measured by Randolph's Kappa~\cite{Randolph:2005}.
    Agreement of models with humans is the average of pairwise model-human agreements.
}
\label{tab:iaa-evaluation}
\vspace{-0.7em}
\end{table}

\begin{figure*}[t]
\centering
\scriptsize
\resizebox{\textwidth}{!}{
\begin{tabular}{p{\textwidth}}
\toprule

\Example
{... Two experiments were conducted. In Experiment 1, COCs (n = 420) were randomly assigned to four groups: ... \hlredmuted{In Experiment 2, COCs (n = 581) were assigned to the same groups, but those in VS1, VS1 + VS2 and Vitrified groups were sub-divided and exposed to the warming solution for either 1 or 5 min.} ... In Experiment 2, there was no effect of time in the warming solution...}
{... We did two experiments. In the first experiment, we had 420 [cow egg cells (also called COCs)] separated into four groups: ... In the second experiment, we treated 581 COCs the same way but also had some of the COCs from groups 2, 3 and 4 spend either 1 or 5 minutes in a warming solution....}
{What were the details of Experiment 2?}
{In Experiment 2, 581 COCs were assigned to the same groups as in Experiment 1, but those in VS1, VS1 + VS2 and Vitrified groups were sub-divided and exposed to the warming solution for either 1 or 5 minutes.}
{\AnswerableXcMulti\ \AnswerableXsVague\ \AccuracyExtPartial\ \AccuracyAbsPartial\ \GivennessGood\ \LocalizationMissing\ \JargonAA\ \StandaloneBad\ }{\Llama}{\textbf{Example 1: Vague question with multiple valid ways of answering; The answer is not standalone because acronym ``VS'' is only defined in $X_{\text{src}}$.}}
\\\hline\addlinespace[0.3em]

\Example
{\hlredmuted{One-hundred participants with a mean±s.d. age of 55.6±10.6 year, body weight of 102.9±18.4 kg and HbA1c of 7.7±1.3\% were randomly} ...}
{We had 100 participants on average about 55 years old, weighing about 102.9 kilos, with a blood sugar level of 7.7\%. They were placed randomly ...}
{What were the ages, weights, and HbA1c levels of the participants in the study?\newline}
{The participants had a mean age of 55.6 years, a mean weight of 102.9 kg, and a mean HbA1c level of 7.7\%.}
{\AnswerableXcGood\ \AnswerableXsYes\ \AccuracyExtCorrect\ \AccuracyAbsCorrect\ \GivennessBad\ \LocalizationMissing\ \JargonA\ \StandaloneGood\ }{\Llama}{\textbf{Example 2: Not an information loss: $A$ does not add any information to $X_{\text{tgt}}$. Givenness: ``HbA1c'' is not known from the simple text or general knowledge.}}
\\\hline\addlinespace[0.3em]

\Example
{... Group L received a \hlorangemuted{loading dose of 0.5 $\mu$g/kg} for 10 minutes while group H received 1.0 $\mu$g/kg...}
{... We divided these patients into two groups. One received a \hlorangemuted{lower dosage of the drug}, while the other received a higher dosage.}
{What was the specific loading dose used in the study?}
{The patients in group H received a loading dose of 1.0 $\mu$g/kg.}
{\AnswerableXcMulti\ \AnswerableXsNo\ \AccuracyExtPartial\ \AccuracyAbsPartial\ \GivennessBad\ \LocalizationGood\ \JargonA\ \StandaloneBad\ }{\Mistral}{\textbf{Example 3: Both evidence and answer are incomplete because they miss the loading dose of one group.}}
\\
\bottomrule
\end{tabular}}
\vspace{-0.5em}
\caption{Qualitative examples demonstrating error cases. More examples in \cref{fig:examples-continued}.}
\label{fig:examples}
\vspace{-0.5em}
\end{figure*}

\section{Related Work}
\paragraph{Deletion in text simplification.}
Professional editors commonly use deletion to make text more accessible~\cite{Petersen:2007:SLaTE,Xu:2015:TACL,Zhong:2020:AAAI,Yamaguchi:2023:EACL}.
\citet{Devaraj:2022:ACL} found that even professional simplifications include \emph{over}-deletions, and that models are prone to mimic this behavior.
In user studies, \citet{Agrawal:2024:TACL} found that deletions are a major factor for diminished reading comprehension.
These studies highlight the importance of detecting and mitigating deletions.
We contribute to this area by providing the first annotated dataset of information loss and a QA-based framework for addressing it.

\paragraph{Question generation (QG).}
While early work considered QG with factoid answers, the focus shifted to more natural, information-seeking and inquisitive questions~\cite{Kwiatkowski:2019:TACL,Ko:2020:EMNLP,Scialom:2020:COLING,Dasigi:2021:NAACL,Meng:2023:IJCNLP-AACL}.
We consider questions with a similar pragmatic goal as clarification questions~\cite{Rao:2018:ACL,Majumder:2021:NAACL} and gap-focused questions in dialogue~\cite{Rabin:2023:ACL}: Asking about information which is missing or vague in a context.
\citet{Newman:2023:EMNLP} demonstrated the merits of QG to represent missing information for decontextualization.
But we draw attention to the particularity of QG in text simplification.
Compared to experts which have expectations of what information texts typically include, lay readers have difficulty asking these clarification questions due to their lacking ``disciplinary knowledge''~\cite{August:2023:TOCHI}, i.e., they have \emph{unknown unknowns}.

Closely related is \textsc{DiffQG}~\cite{Cole:2023:EACL} which uses QG to describe \emph{factual} changes in two revisions of a Wikipedia passage.
In contrast, we consider simplification-induced changes where answers to questions are not necessarily different but vague.
Also, our task requires document-level comparisons, simple factoid to complex multi-sentence answers, and tailoring the QA to laypeople.

\paragraph{QA for evaluation and entailment.}
QA has been used to evaluate summarization and simplification~\cite{Mani:2002:NLE,Agrawal:2024:TACL}.
The advances in QG gave rise to the cross-questioning paradigm~\citep[][\emph{inter alia}]{Wang:2020:ACL,Durmus:2020:ACL,Deutsch:2021:TACL}.
This line of work differs from \textsc{InfoLossQA} in two important aspects.
First, there is a strong focus on noun-phrase and entity-centric QA.
Our proposed task invites QAs that are meant for consumption by end-users, and hence must satisfy additional linguistic criteria (\cref{sec:task}).
Second, while its conceivable to use cross-questioning to get candidate QAs, \citet{Kamoi:2023:EACL} identified error-propagation in the QG stage as a fundamental limitation making QA-level answerability unreliable.
They therefore advocate for NLI-based approaches~\cite{Laban:2022:TACL,Rodriguez:2023:arXiv}, which our pipeline-system is inspired by.

\section{Conclusion}
We propose \textsc{InfoLossQA}: A task and dataset to describe and recover simplification-induced information loss as reader-centric QAs.
Our analyses show that QAs following the QUD theory provide a rich description that can mitigate overdeletion and vagueness in text simplification.
We also establish automatic pipelines for the task and propose a rigorous evaluation framework considering correctness of QAs and user-centric constraints.
Looking ahead, this work opens new avenues in interactive simplification tools~\cite{Fok:2023:arXiv} and for quality assessment of automatic simplifications.

\section*{Limitations}
Our motivation and vision for \textsc{InfoLossQA} is to help users to deepen their understanding of a text.
However, we focus on the technical dimensions of this goal: establishing a dataset, an evaluation framework, and developing and evaluating baselines.
Testing the effects of the proposed QA on \emph{end-user} comprehension is an important direction for future work in interactive text simplification.

Furthermore, while we believe that the proposed approach is generalizable, our experiments are confined to one language (English), text genre (abstracts of medical publications) and simplification style (GPT-4 simplification).
Future work could assess the versatility of this framework under different conditions.
For instance, we considered simplifications at one level of compression, but readers may prefer different degrees of simplification~\cite{Xu:2015:TACL,August:2024:CHI}.
As we alter the degree of simplification, it becomes important to understand (a) how effective models are at identifying information loss, and (b) how the distribution of information loss and associated QAs changes.

Lastly, our proposed evaluation framework currently relies on human annotators to judge the quality of model outputs.
Any kind of human evaluation comes at a significant cost which may limit further studies on better modeling.
Therefore, future work could develop automatic metrics for the task.
Our initial experiments show that automatic LLM-based evaluation is a promising method, but requires additional work to achieve good correlations with human judgments.
To facilitate this direction, we release the full evaluation suite including human judgments and evaluation baselines.

\section*{Acknowledgments}
We thank Keziah Kaylyn Reina, Kathryn Kazanas and Karim Villaescusa F. for their annotation and evaluation effort, David Heinemann for help with the annotation interface, and Ritvik Renikunta for help with the initial design of the end-to-end prompts.
We also thank Eunsol Choi for feedback on this paper, and Juan Diego Rodriguez and Manya Wadhwa for useful discussions.
This research is partially supported by NSF CAREER Award IIS-2145479 and Good Systems,\footnote{\url{https://goodsystems.utexas.edu}} a UT Austin Grand Challenge to develop responsible AI technologies.
Trienes was supported by the Cancer Research Center Cologne Essen (CCCE), the Federal Ministry of Education and Research (BMBF) and by a fellowship within the IFI programme of the German Academic Exchange Service (DAAD).
Wallace was supported in this work by the National Institutes of Health (NIH), grant R01LM012086, and by the National Science Foundation (NSF), grant 1750978. Xu is supported in part by NSF awards IIS-2144493 and IIS-2112633.

\bibliography{bibliography}

\appendix
\counterwithin{figure}{section}
\counterwithin{table}{section}
\section{Appendix}

\subsection{Analysis: Relation to Paper Plain Key Question Index}
\label{sec:appendix-faqs}
In addition, to the domain-agnostic question taxonomy~\cite{Cao:2021:ACL} used in \cref{sec:analysis}, we analyzed the QA pairs through the lens of the Paper Plain Key Question Index which was designed to convey the most important elements of an RCT~\cite{August:2023:TOCHI}.
We manually codify 120 QA pairs (40 per annotator) according to the 8 categories of the question index.

Aligned with findings in \cref{sec:analysis}, the information which is most likely lost is about \emph{methodological detail} (i.e., study protocol, analysis tools, population; 53\% of QA pairs), followed by \emph{results} (18\% of QA pairs) and the \emph{goals} of the trial (11\% of QA pairs).
Critical information like the usual and new treatments is preserved.
Most notably, we rarely observe outright omission of \emph{all} information regarding a top-level question in the key question index. Instead, the \textsc{InfoLossQA} questions cover information with a higher level of specificity.
See \cref{tab:appendix-paperplain-faq} for a detailed breakdown and examples.

\subsection{Analysis: How are Errors Distributed Across Document Sections?}
\begin{figure*}[t]
\centering
\includegraphics[width=\textwidth]{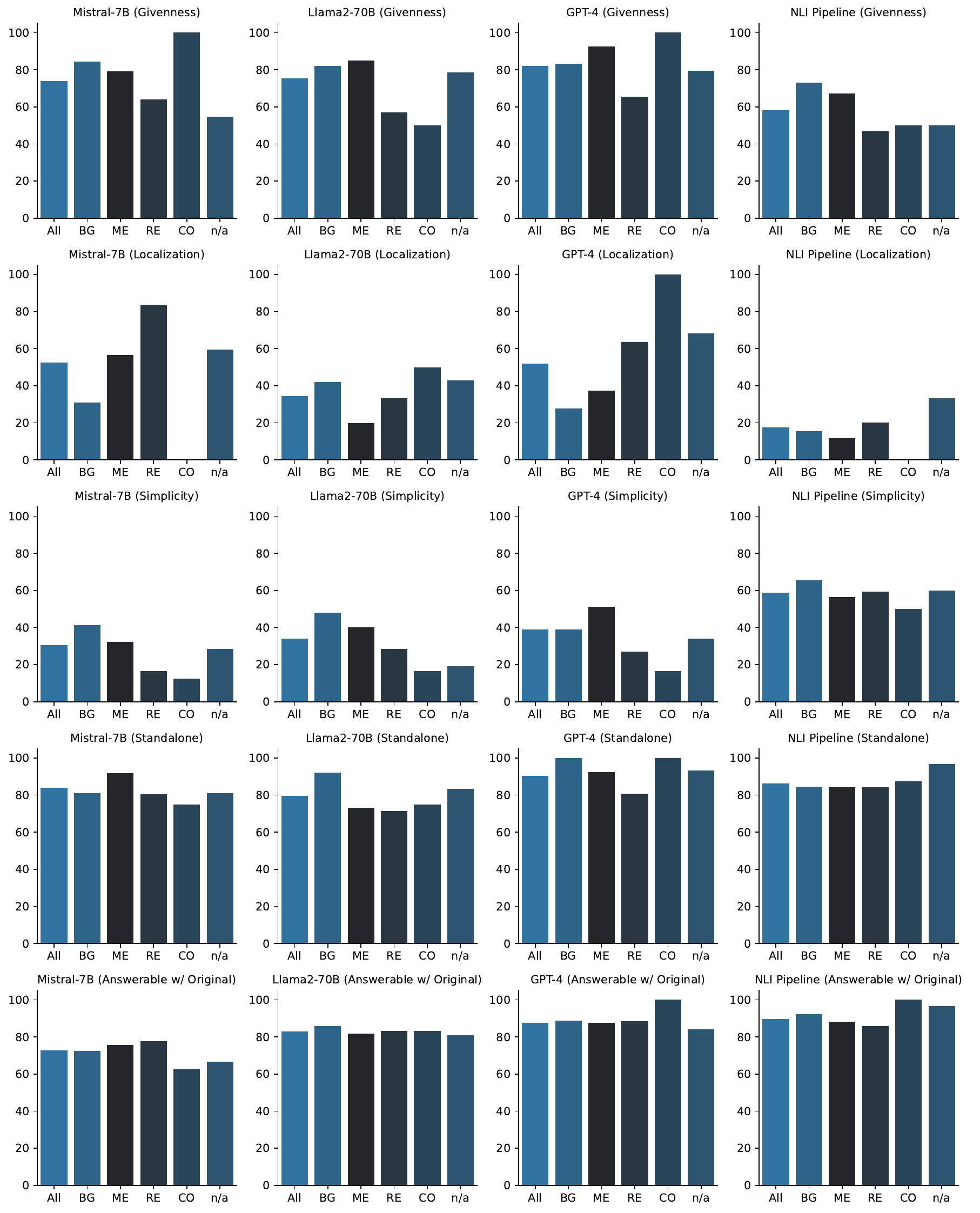}
\caption{Percentage of QAs with a good rating grouped by model, evaluation criterion and RCT section (BG = Background; ME = Methods; RE = Results; CO = Conclusion; n/a for RCTs without sections). Criteria with multiple response options were binarized to ``good'' vs. ``not good.'' Continued in~\cref{fig:eval-by-section-b}}
\label{fig:eval-by-section}
\end{figure*}

Given that some RCT sections are more technical than others, it is conceivable that models are better generating information loss QAs in some sections than in others.
We plot the percentage of good responses per model, section and evaluation criterion in \cref{fig:eval-by-section}.
Contrary to our expectation, there are no notable differences across sections, with only a slight trend for reduced Givenness and Jargon for QAs localized in the results and conclusion section.
Qualitatively, this is often due to questions asking about the statistical significance of the results, which annotators deemed to be an unfamiliar concept for lay readers.

\subsection{Experiment Detail: Manual Evaluation}
\label{sec:appendix-evaluation}
\paragraph{Sampling.}
For aspect 1+2, we take a stratified sample to preserve the relative frequencies at which models generate QAs per RCT section. As the generations by Mistral-7B and Llama2-70B do not always follow the specified output format, we only sample from QA pairs which could be completely parsed. See \cref{tab-generation-errors} for a detailed analysis of generation errors.

\paragraph{Recall evaluation: Finding candidate matches.}
For our dataset of 1,000 reference QAs across 104 documents and predictions by four models, the pairwise recall evaluation described in \cref{sec:evaluation} results in 33,825 comparisons which is intractable.
Therefore, we follow the paraphrase mining approach by~\citet{Wright:2022:EMNLP} to get \emph{candidate matches}.
First, we calculate the cosine similarity of two QA pairs using sentence embeddings~\cite{Reimers:2019:EMNLP}. Both the question and the answer are concatenated before calculating the embedding.
Afterwards, we establish a threshold by annotating 400 matches (predicted QA and reference QA) equally sampled from 20 bins in the similarity range of $[0, 1]$.
We observe the first bin with a notable number of recalled QA (here: $T = 0.65$).
Pairs with similarity $\leq T$ are set to \emph{not recalled}.
This process reduced the required manual evaluations by 93\% to 2,466.

We aggregate votes as follows. A reference QA is set to \emph{fully recalled} if at least one predicted QA fully recalls it. If a reference QA is not fully recalled but partially recalled by at least one predicted QA, it is set to \emph{partially recalled}. Otherwise it is set to \emph{not recalled}

\onecolumn
\begin{table}[H]
\centering
\small
\begin{tabular}{p{.4\linewidth}lp{.3\linewidth}}
\toprule
\textbf{Description} & \textbf{Prompt} & \textbf{Decoding Parameters} \\
\midrule
\emph{Simplification.} Simplifying technical RCT abstracts ($X_{\text{src}}$) into a plain language version ($X_{\text{tgt}}$). See top part of \cref{fig:compositionality-example} for an example simplification. &
\cref{fig:prompt-simplification} &
\texttt{\small model=gpt-4-0613}\newline
\texttt{\small temperature=1.0}\newline
\texttt{\small max\_tokens=1024}\newline
\texttt{\small top\_p=1}\newline
\texttt{\small frequency\_penalty=0}\newline
\texttt{\small presence\_penalty=0}
\\\midrule
\emph{Few-shot question classifier.} Categorizing questions according to the typology of \citet{Cao:2021:ACL}. This classifier obtains an F1 score of 0.87 on a manually labeled validation set of 50 questions in \textsc{InfoLossQA}. &
\cref{fig:gpt4-prompt-question-classifier} &
\texttt{\small model=gpt-4-0613}\newline
\texttt{\small temperature=0}\newline
\texttt{\small max\_tokens=1024}\newline
\texttt{\small top\_p=1}\newline
\texttt{\small frequency\_penalty=0}\newline
\texttt{\small presence\_penalty=0}
\\\midrule
\multicolumn{3}{p{.85\linewidth}}{\textbf{NLI pipeline for information loss detection.}}
\\\midrule
\emph{Part 1: fact extraction.} Extracting atomic facts from $X_{\text{src}}$. These facts are classified for entailment with $X_{\text{tgt}}$ with a standard NLI model~\cite{Laurer:2024}. &
\cref{fig:gpt4-prompt-fact-extraction} &
\texttt{\small model=gpt-4-0613}\newline
\texttt{\small temperature=0}\newline
\texttt{\small max\_tokens=512}\newline
\texttt{\small top\_p=1}\newline
\texttt{\small frequency\_penalty=0}\newline
\texttt{\small presence\_penalty=0}
\\\midrule
\emph{Part 2: QA-generation.} Generating QA pairs based on $X_{\text{src}}$, $X_{\text{tgt}}$ and the list of facts with \emph{neutral} outcome of the NLI classifier. &
\cref{fig:gpt4-prompt-question-generation} &
\texttt{\small model=gpt-4-0613}\newline
\texttt{\small temperature=1}\newline
\texttt{\small max\_tokens=4096}\newline
\texttt{\small top\_p=1}\newline
\texttt{\small frequency\_penalty=0}\newline
\texttt{\small presence\_penalty=0}
\\\midrule
\multicolumn{3}{p{.85\linewidth}}{
\textbf{End-to-end prompts for information loss detection.} All prompts include a one-shot example, which we found to substantially improve performance over a zero-shot prompt. For a fair comparison across models, we do not include more than one example as it would exhaust the context windows of some models under investigation.}
\\\midrule
\emph{GPT-4}~\cite{OpenAI:2023:arXiv}. End-to-end prompt to detect information loss and generate QA pairs. Inference on API of OpenAI. &
\cref{fig:gpt4-prompt} &
\texttt{\small model=gpt-4-0613}\newline
\texttt{\small temperature=0}\newline
\texttt{\small max\_tokens=2048}\newline
\texttt{\small top\_p=1}\newline
\texttt{\small frequency\_penalty=0}\newline
\texttt{\small presence\_penalty=0}
\\\midrule
\emph{Mistral-7B}~\cite{Jiang:2023:arXiv}. End-to-end prompt to detect information loss and generate QA pairs. Inference with the Huggingface transformers library~\cite{Wolf:2020:EMNLP} on one NVIDIA RTX A6000 (48GB) completed in less than one hour. &
\cref{fig:mistral-prompt} &
\texttt{\small model=Mistral-7B-Instruct-v0.1}\newline
\texttt{\small do\_sample=False}\newline
\texttt{\small temperature=0}\newline
\texttt{\small max\_tokens=2048}\newline
\texttt{\small top\_p=1}\newline
\texttt{\small top\_k=1}\newline
\texttt{\small repetition\_penalty=1}
\\\midrule
\emph{Llama2-70B}~\cite{Touvron:2023:arXiv}. End-to-end prompt to detect information loss and generate QA pairs. Inference on API of \rurl{Together.AI}. &
\cref{fig:llama2-prompt} &
\texttt{\small model=llama-2-70b-chat}\newline
\texttt{\small max\_tokens=None}\newline
\texttt{\small temperature=0}\newline
\texttt{\small top\_p=1}\newline
\texttt{\small top\_k=1}\newline
\texttt{\small repetition\_penalty=1}
\\
\bottomrule
\end{tabular}
\caption{LLM prompts and decoding parameters.}
\label{tab:prompts-overview}
\end{table}

\begin{table*}[t]
\small
\setlength{\tabcolsep}{5pt}
\centering
\begin{tabular}{lrrrrrrrrr}
\toprule
\textbf{Annotator / Model} & Docs. & QA & \% Over & \% Del & QA/doc & $|Q|$ & $|A|$ & $|E|$ & $|R|$ \\
\midrule
Annotator 1 & 48 & 192 & 90.1 & 9.9 & 4.1 & 10.8 & 23.8 & 12.2 & 7.6 \\
Annotator 2 & 75 & 425 & 71.8 & 28.2 & 5.7 & 12.2 & 33.4 & 15.5 & 9.0 \\
Annotator 3 & 85 & 383 & 68.7 & 31.3 & 4.7 & 10.9 & 30.2 & 14.4 & 8.6 \\
\midrule
All (micro avg.) & 104 & 1000 & 74.1 & 25.9 & 9.6 & 11.4 & 30.3 & 14.4 & 8.5 \\
\addlinespace
Mistral-7B & 104 & 507 & 45.2 & 54.8 & 4.9 & 11.6 & 26.2 & 22.3 & 18.1 \\
Llama2-70B & 104 & 681 & 38.2 & 61.8 & 6.7 & 12.9 & 30.6 & 20.4 & 15.7 \\
GPT-4 & 104 & 477 & 48.4 & 51.6 & 4.6 & 14.4 & 33.7 & 25.4 & 20.6 \\
NLI Pipeline & 104 & 1699 & -- & 100.0 & 16.5 & 14.5 & 24.6 & 34.7 & -- \\
\bottomrule
\end{tabular}
\caption{Summary statistics of human-written QAs and model predictions. Over = Oversimplification; Del = Deletion; Length of question $Q$, answer $A$, evidence spans $E \in X_{\text{src}}$ and rationale spans $R \in X_{\text{tgt}}$ is given in tokens.}
\label{tab:stats-breakdown}
\end{table*}

\begin{table*}
\small
\centering
\begin{tabular}{lrrrr}
\toprule
\textbf{Error} & \textbf{Mistral-7B} & \textbf{Llama2-70B} & \textbf{GPT-4} & \textbf{NLI Pipeline} \\
\midrule
\hlred{Deletion} (Total) & 278 & 421 & 246 & 1699 \\\midrule
\% Valid (no error) & 61.9 & 87.2 & 100 & 100 \\
\% Spurious $R$ & 38.1 & 1.0 & - & - \\
\% Invalid $E$ & - & 11.9 & - & - \\\midrule
\hlorange{Oversimplification} (Total) & 229 & 260 & 231 & 0 \\\midrule
\% Valid (no error) & 67.2 & 78.8 & 100 & - \\
\% Invalid $R$ & 28.8 & 5.8 & - & - \\
\% Missing $R$ & 2.2 & 6.2 & - & - \\
\% Invalid $E$ & 1.7 & 2.3 & - & - \\
\% Invalid $E$ + Invalid $R$ & - & 5.8 & - & - \\
\% Invalid $E$ + Missing $R$ & - & 1.2 & - & - \\
\bottomrule
\end{tabular}
\caption{Analysis of LLM generation errors grouped by deletion and oversimplification. For each category, the total number of QA pairs is given with the percent of QAs per error. Spurious $R$: QAs classified as deletions should not have a rationale span. Invalid $E$/$R$: the span cited by the model is not a valid substring of $X_{\text{src}}$/$X_{\text{tgt}}$.}
\label{tab-generation-errors}
\end{table*}

\begin{table*}[t]
\small
\centering
\resizebox{\textwidth}{!}{%
\begin{tabular}{lrp{.7\textwidth}}
\toprule
\textbf{QA Category} & \textbf{\%} & \textbf{Definition and Examples} \\
\midrule
Motivation\textsuperscript{\textdagger} & 5.0\% & \textbf{Why was the study conducted?}
\newline Q: What was the motivation for this study to investigate the dosage of haloperidol to address nausea and vomiting after surgery?
\newline A: There is evidence that a small dose of haloperidol can help prevent...
\\\addlinespace[0.15em]\hline\addlinespace[0.25em]

Condition & 5.8\% & \textbf{What condition does this paper study?} \\

-- Inclusion Criteria\textsuperscript{\textdagger} & 3.3\% & \emph{What were the specific inclusion criteria for participants?}
\newline Q: What criteria did the researchers use to select eligible participants?
\newline A: The researchers gathered participants between the ages of 3-18, ...
\\\addlinespace

-- Other\textsuperscript{\textdagger} & 2.5\% & \emph{Other condition-related question.}
\newline Q: What kind of hip surgery did the participants receive?
\newline A: The participants were getting hip surgery under the subarachnoid block...
\\\addlinespace[0.15em]\hline\addlinespace[0.25em]

Goal & 10.8\% & \textbf{What did the paper want to find out?}
\newline Q: What about the nicotine vaccine could potentially help smokers quit?
\newline A: The nicotine vaccine boosts antibody concentrations, which helps to stimulate...
\\\addlinespace[0.15em]\hline\addlinespace[0.25em]

Usual Treatment & 1.7\% & \textbf{How is the condition usually treated?}
\newline Q: What type of active deep brain stimulation was used in previous trials?
\newline A: Previous trials found a positive impact of unilateral (only applied to one side of the brain) active deep brain stimulation on symptoms of Parkinson's disease.
\\\addlinespace[0.15em]\hline\addlinespace[0.25em]

New Treatment & 4.2\% & \textbf{What were the new treatment(s), if any this paper looked into?}
\newline Q: What type of formoterol is being observed in the study?
\newline A: Long-acting beta2-agonist formoterol, which is a drug used to treat asthma...
\\\addlinespace[0.15em]\hline\addlinespace[0.25em]

Method & 53.3\% & \textbf{What did the paper do?} \\

-- Outcome (Tool)\textsuperscript{\textdagger} & 14.2\% & \emph{What tools/procedures were used to measure the effects of interventions?}
\newline Q: How was sleep quality and life quality measured for participants?
\newline A: Sleep quality and life quality were assessed using the Sleep-Apnoea-Quality-of-Life-Index...
\\\addlinespace

-- Study Protocol\textsuperscript{\textdagger} & 10.8\% & \emph{What was the protocol of the study?}
\newline Q: What was the setting and design of the study? %
\newline A: The study was an open (both participants and researchers knew who was assigned which drops), cross-over (the participants receive both treatments ...), comparative study (comparing the two drops).
\\\addlinespace

-- Quantity\textsuperscript{\textdagger} & 7.5\% & \emph{With what dosage/quantity/frequency were the interventions performed?}
\newline Q: How much lurasidone was given to the patients in the trial?
\newline A: Patients were given 40 to 80 milligrams of lurasidone per day, given flexibly.
\\\addlinespace

-- Population\textsuperscript{\textdagger} & 7.5\% & \emph{What were the demographics of the patients in the study?}
\newline Q: What was the age range of the patients?
\newline A: 52-85 years.
\\\addlinespace

-- Analysis Technique\textsuperscript{\textdagger} & 5.8\% & \emph{How were the results analyzed?}
\newline Q: How were the results analyzed?
\newline A: The results of the test were analyzed using a t-test and... 
\\\addlinespace

-- Other\textsuperscript{\textdagger} & 7.5\% & \emph{Other method-related question.}
\newline Q: How often were participants checked during the study?
\newline A: Participants were assessed on their first visit, on the 7th day or their first follow-up, and on the 14th day or their second follow-up.
\\\addlinespace[0.15em]\hline\addlinespace[0.25em]

Results & 18.3\% & \textbf{What did the paper find?}
\newline Q: How did the number of evens in the Tamoxifen group compare to the other group?
\newline A: The Tamoxifen group had a total of 39 events, while the other...
\\\addlinespace[0.15em]\hline\addlinespace[0.25em]

Contrasting Results & 0.8\% & \textbf{Are the findings different depending on a person’s demographics?}
\newline Q: How did selenium levels of pregnant women in the UK vary from those of pregnant women in other countries?
\newline A: The selenium (mineral) levels of pregnant women in the UK were lower compared to those of pregnant women in other countries...
\\
\bottomrule
\end{tabular}}
\caption{Classification of 120 randomly sampled human QA pairs in \textsc{InfoLossQA} according to the pragmatic goals of questions in the Paper Plain Key Question Index developed for lay readers of RCTs~\cite{August:2023:TOCHI}. Frequency of questions is given in percent (\%). (Sub-)Categories marked with \textsuperscript{\textdagger} were added by us to give a more fine-grained view on the questions types.}
\label{tab:appendix-paperplain-faq}
\end{table*}

\begin{figure*}[t]
\centering
\scriptsize
\resizebox{\textwidth}{!}{
\begin{tabular}{p{\textwidth}}
\toprule

\Example
{[The aim of this study is] to compare the efficacy of high and low oral doses of haloperidol in the treatment of \hlorangemuted{acute and transient psychotic disorders}...}
{This study is about testing whether high or low doses of a drug called haloperidol are more effective at treating \hlorangemuted{short-lasting mental illnesses}...}
{What specific type of mental illness were the patients suffering from?}
{The patients were suffering from acute and transient psychotic disorders, which are characterized by sudden onset of psychotic symptoms that last for a short period of time.}
{\AnswerableXcGood\ \AnswerableXsVague\ \AccuracyExtCorrect\ \AccuracyAbsCorrect\ \GivennessGood\ \LocalizationGood\ \JargonAAA\ \StandaloneGood\ }{\GPT}{\textbf{Example 4: Good generation.}}
\\\hline\addlinespace[0.3em]

\Example
{Since the prognosis of advanced biliary tract cancer (aBTC) still remains very poor... \hlredmuted{Patients with aBTC refractory to at least one regimen of chemotherapies} were randomly assigned to receive PPV with low dose CPA ... (PPV/CPA, n = 24) or PPV alone (n = 25).}
{Biliary tract cancer (aBTC), a type of liver cancer, has a very poor survival rate. To help improve this, we're researching new treatments ... [W]e gave it to 24 patients and compared them to a group of 25 patients who received only the PPV.}
{Who were the participants in the study?\newline}
{The participants were patients with advanced biliary tract cancer (aBTC) who had failed to respond to at least one regimen of chemotherapies.}
{\AnswerableXcGood\ \AnswerableXsNo\ \AccuracyExtCorrect\ \AccuracyAbsCorrect\ \GivennessGood\ \LocalizationMissing\ \JargonAAAA\ \StandaloneGood\ }{\Mistral}{\textbf{Example 5: Missing rationale $R$: annotator pointed out that ``patients'' is an oversimplification of ``patients with aBTC,'' so it should be highlighted.}}
\\\hline\addlinespace[0.3em]

\Example
{\hlredmuted{This study investigated the effect of combining oral dexamethasone with either nebulized racemic epinephrine or salbutamol compared to bronchodilators alone for the treatment of infants with bronchiolitis.} ... This was a double-blind, randomized controlled trial on infants (1 to 12 months) who were diagnosed in the emergency department with moderate-to-severe bronchiolitis...}
{This research looked at how well infants with a common lung infection called bronchiolitis responded to different combinations of medicines... The study investigated the effects on babies between 1 to 12 months old...}
{Where were the infants diagnosed with bronchiolitis?}
{The infants were diagnosed in the emergency department of a hospital.}
{\AnswerableXcGood\ \AnswerableXsNo\ \AccuracyExtWrong\ \AccuracyAbsCorrect\ \GivennessGood\ \LocalizationCorrectlyAbsent\ \JargonAAAA\ \StandaloneGood\ }{\NLI}{\textbf{Example 6: Even though the QA correctly reveals missing information, the highlighted evidence $E$ does not support the answer.}}
\\
\bottomrule
\end{tabular}}
\caption{Qualitative examples demonstrating error cases. Continued from \cref{fig:examples}.}
\label{fig:examples-continued}
\end{figure*}

\begin{figure*}[t]
\centering
\includegraphics[width=\textwidth]{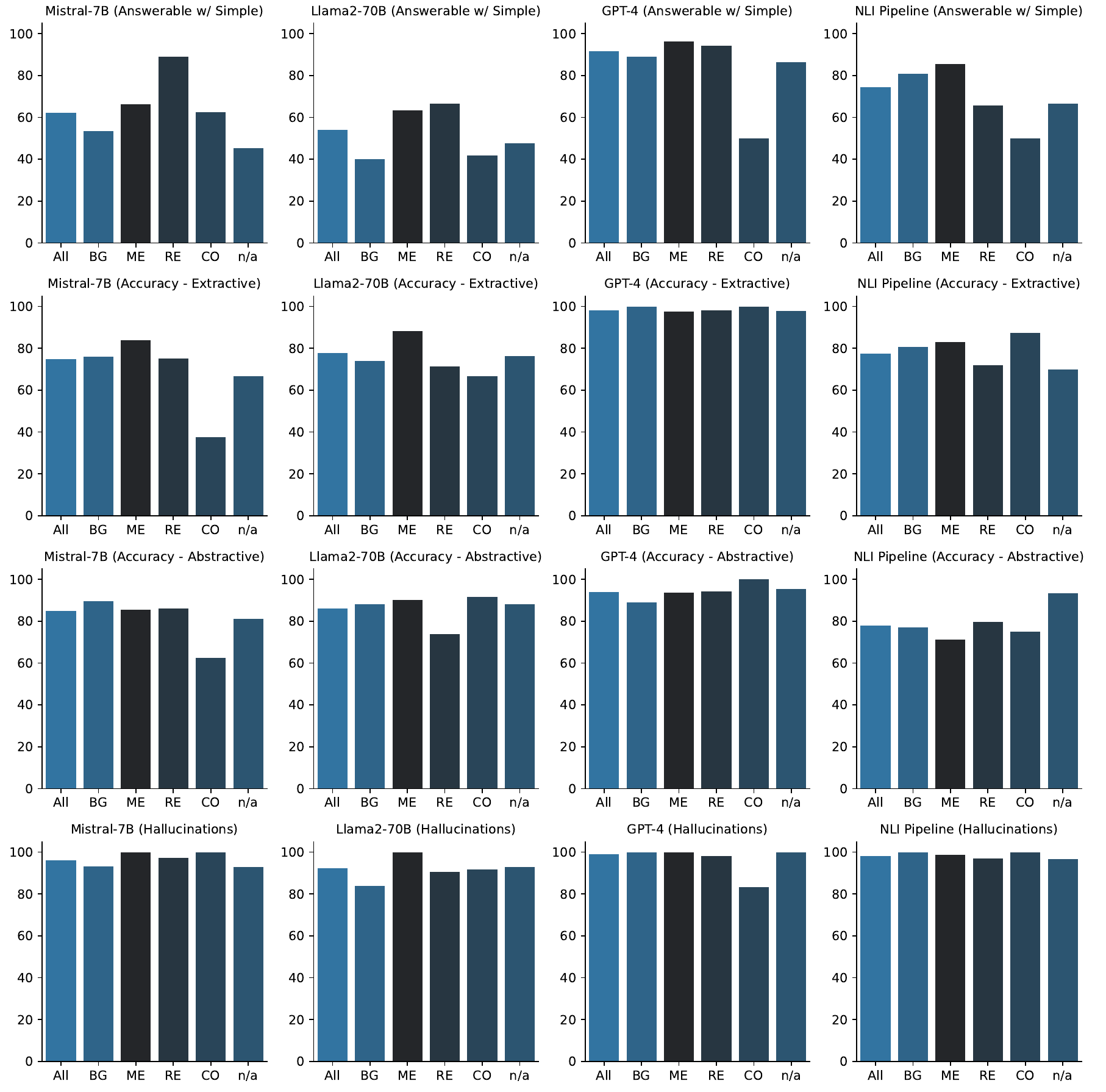}
\caption{Percentage of QAs with a good rating grouped by model, evaluation criterion and RCT section (BG = Background; ME = Methods; RE = Results; CO = Conclusion; n/a for RCTs without sections). Criteria with multiple response options were binarized to ``good'' vs. ``not good.'' Continued from~\cref{fig:eval-by-section}}
\label{fig:eval-by-section-b}
\end{figure*}

\begin{figure*}[t]
\centering
\includegraphics[width=\textwidth]{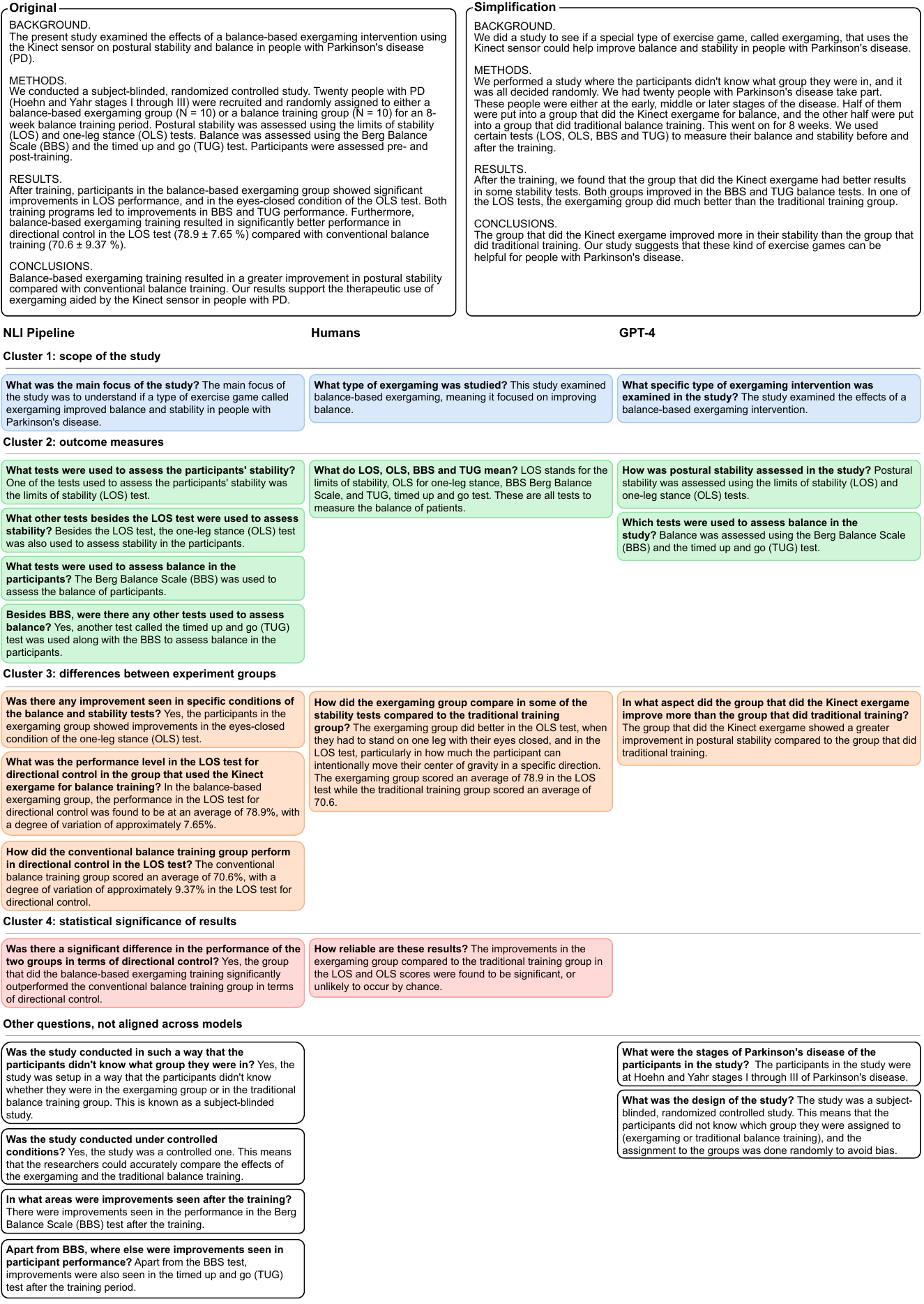}
\caption{Full example comparing QAs generated by the NLI pipeline, humans and GPT-4. Colors indicate semantic alignment between questions. We observe different tendencies for question compositionality: while humans aggregate several missing information into higher-level questions, the NLI pipeline has a tendency to generate more restricted questions, even though in aggregate they cover a similar content.}
\label{fig:compositionality-example}
\end{figure*}


\begin{figure*}
\scriptsize
\begin{tcolorbox}[colback=pastelgreen!40, colframe=pastelgreen, arc=1mm, boxrule=1pt, left=2mm, right=2mm, top=0mm, bottom=0mm]
\begin{lstlisting}
Please simplify the following technical abstract into plain language that an average adult would understand. If the abstract has sections, keep them.
\end{lstlisting}
\end{tcolorbox}
\begin{tcolorbox}[colback=pastelorange!40, colframe=pastelorange, arc=1mm, boxrule=1pt, left=2mm, right=2mm, top=0mm, bottom=0mm]
\begin{lstlisting}
{{ abstract }}
\end{lstlisting}
\end{tcolorbox}
\caption{GPT-4 prompt to simplify the RCT abstracts. Colors indicate chat roles: \hlgreen{system} and \hlorange{user}.}
\label{fig:prompt-simplification}
\end{figure*}

\begin{figure*}
\scriptsize
\begin{tcolorbox}[colback=pastelgreen!40, colframe=pastelgreen, arc=1mm, boxrule=1pt, left=2mm, right=2mm, top=0mm, bottom=0mm]
\begin{lstlisting}
You are a helpful assistant to classify text into categories.

## Instructions
You are asked to classify questions according to an ontology of question types. The question type reflects the nature of the question. It is NOT determined by the interrogative word of the question. There are 10 question types in total. The definition for each type is shown below. Please select the question type which is most likely for a given question. Only output the category title, not the description.

## Question Types
1. VERIFICATION: Asking for the truthfulness of an event or a concept.
- Was the study double-blinded?
- Was there a trend towards smaller increases in macular pigment for subjects with high baseline values?

2. DISJUNCTION: Asking for the true one given multiple events or concepts, where comparison among options is not needed.
- no example available, match by the description

3. CONCEPT: Asking for a definition of an event or a concept.
- What kind of result is being measured in this study?
- What were the main inclusion criteria for this study?
- What does the WCJ-III test specifically measure?
- Which areas of the brain were analyzed?

4. EXTENT: Asking for the extent or quantity of an event or a concept.
- How reliable are these results?
- How long were the participants observed?
- How much ibuprofen was in the small dose?
- How many young and old people participated in the study?

5. EXAMPLE: Asking for example(s) or instance(s) of an event or a concept.
- What kind of conditions cause corneal neovascularization?

6. COMPARISON: Asking for comparison among multiple events or concepts.
- On what results did the control group do better than the intervention group?
- How did headache of participants receiving ibuprofen compare to those participants that received a placebo?
- How more effective was the arm cranking exercise with and without electrical muscle stimulation?

7. CAUSE: Asking for the cause or reason for an event or a concept.
- What motivates this study?
- Why is EMS being investigated?

8. CONSEQUENCE: Asking for the consequences or results of an event.
- What was the effect of using ibuprofen to treat headaches?
- What were the main findings of the study?
- What does this study tell us about arm cranking with electrical muscle stimulation?

9. PROCEDURAL: Asking for the procedures, tools, or methods by which a certain outcome is achieved.
- What questionnaire was used for this study?
- What kind of lab tests were done?
- How were the patients assigned to a group?
- How were the different drugs administered to participants?

10. JUDGMENTAL: Asking for the opinions of the answerer's own.
- no example available, match by the description
\end{lstlisting}
\end{tcolorbox}
\begin{tcolorbox}[colback=pastelorange!40, colframe=pastelorange, arc=1mm, boxrule=1pt, left=2mm, right=2mm, top=0mm, bottom=0mm]
\begin{lstlisting}
## Instances to classify
{% for i, question in enumerate(questions) %}
    {{ i+1 }}. {{ question }}
{% endfor %}
\end{lstlisting}
\end{tcolorbox}
\caption{Few-shot prompt for GPT-4 to classify questions according to the taxonomy of~\citet{Cao:2021:ACL}. Colors indicate chat roles: \hlgreen{system} and \hlorange{user}.}
\label{fig:gpt4-prompt-question-classifier}
\end{figure*}

\begin{figure*}
\scriptsize
\begin{tcolorbox}[colback=pastelgreen!40, colframe=pastelgreen, arc=1mm, boxrule=1pt, left=2mm, right=2mm, top=0mm, bottom=0mm]
Statement: \{\{ sent \}\}

List all the facts we explicitly know from the statement. Make each fact as atomic as possible.
\end{tcolorbox}
\caption{GPT-4 prompt to decompose a sentence into a list of atomic facts.}
\label{fig:gpt4-prompt-fact-extraction}
\end{figure*}

\begin{figure*}
\scriptsize
\begin{tcolorbox}[colback=pastelgreen!40, colframe=pastelgreen, arc=1mm, boxrule=1pt, left=2mm, right=2mm, top=0mm, bottom=0mm]
\begin{lstlisting}
## Original
{{ original }}

## Simplification
{{ simplification }}

## Missing facts
{{ facts }}

The above facts are missing from the simplified text ("Simplification"). For each fact, please write a question-answer pair that would elicit the missing information from the original text ("Original"). Phrase the question in such a way that a reader can understand it without having seen the original text. It should only contain concepts (entities, events, or states) that were mentioned in the simple text, or concepts that have not been directly mentioned but are generally known or inferable from mentioned ones. The answer should be understandable by an average adult, so please explain technical jargon if necessary. Make each question-answer pair as specific as possible and make sure that they are independent of each other. Ask only about one information unit at a time. Do this for all facts, and format your output as follows:

- Fact:
- Question:
- Answer:
\end{lstlisting}
\end{tcolorbox}
\caption{GPT-4 prompt to generate an \textsc{InfoLossQA} pair given original, simplification and a missing fact.}
\label{fig:gpt4-prompt-question-generation}
\end{figure*}

\begin{figure*}
\scriptsize
\begin{tcolorbox}[colback=pastelgreen!40, colframe=pastelgreen, arc=1mm, boxrule=1pt, left=2mm, right=2mm, top=0mm, bottom=0mm]
\begin{lstlisting}
You are an expert annotator for outputs of text simplification systems. This annotation task is to identify pieces of information that were lost in the simplification process. You will be given two texts: the original and a simplification. Assume that a lay reader only sees the simplification. Identify all things which the reader can NOT learn from the simplification but that they could have learnt from the original.

Please classify each unit of information into one of the following two categories:
1. Omissions: Pieces of information which were not included in the simplification.
2. Imprecise/fuzzy concepts: Pieces of information which are included in the simplification, but that have been simplified to the extent that they became imprecise or completely lost their original meaning.

Afterwards, please write a question-answer pair that would elicit the missing information from the original text. Phrase the question in such a way that a reader can understand it without having seen the original text. It should only contain concepts (entities, events, or states) that were mentioned in the simple text, or concepts that have not been directly mentioned but are generally known or inferable from mentioned ones. The answer should be understandable by an average adult, so please explain technical jargon if necessary. Make each question-answer pair as specific as possible and make sure that they are independent of each other. Ask only about one information unit at a time.

A useful heuristic to decide between omissions and imprecise/fuzzy concepts is to see if the question-answer pair clarifies/expands some topic which is discussed in the simplification. If so, classify it as imprecise/fuzzy concepts, otherwise omission.

Adhere to this output format:
- Original: "<span in original text>"
- Rationale: <a short rationale that describes what makes this an information loss>
- Question: <the question that elicits missing information>
- Answer: <the answer that provides the missing information from the original span>

For fuzzy/imprecise concepts, please also indicate the corresponding span in the simplified text.

Here is an example.

## Original
This study evaluates the effects of vitamin D3 on disease activity and quality of life in ulcerative colitis (UC) patients with hypovitaminosis D. The study was a prospective double-blinded, randomized trial conducted at Community Regional Medical Center, Fresno, CA from 2012-2013. Patients with UC and a serum 25(OH)D level <30 ng/ml were eligible for the study. Enrolled subjects were randomized to receive either 2,000 lU or 4,000 IU of oral vitamin D3 daily for a total of 90 days. The Short IBD Questionnaire (SIBDQ) for quality of life, the Partial Mayo Score for UC disease activity and serum lab tests were compared between the two treatment groups.

## Simplification
This study looks at whether taking vitamin D3 can affect a particular form of bowel disease (ulcerative colitis) and improve the lives of patients with low levels of vitamin D. The study was carried out at a medical center in Fresno, California, between 2012 and 2013. Patients with this disease and low vitamin D levels were included. Participants were randomly given either 2,000 IU or 4,000 IU of oral vitamin D3 daily for 90 days. Researchers used a survey about participant's quality of life and conducted lab tests between the two groups.

## Omissions
- Original: "Partial Mayo Score for UC disease activity"
- Rationale: the simplification does not mention that the Partial Mayo Score was used.
- Question: Besides quality of life survey and the serum tests, what other measures did the researchers use?
- Answer: The researchers also tracked the partial mayo for UC diseases. UC stands for ulcerative colitis and is an inflammatory bowel disease. The mayo score indicates how active or severe the disease is.

## Imprecise/fuzzy concepts
- Original: "The study was a prospective double-blinded, randomized trial"
- Simplification: "The study"
- Rationale: The simplification does not explain the design of the study, it only mentions that it was a "study".
- Question: How did the study control for bias?
- Answer: The study was double-blinded, so that neither the researcher nor the participants knew which treatment each participant received, and it was randomized, meaning the participants were randomly assigned one of the treatments.

- Original: "a serum 25(OH)D level <30 ng/ml"
- Simplification: "low vitamin D levels"
- Rationale: the simplification does not explain the inclusion criteria, namely how low the vitamin D levels of eligible patients were.
- Question: How low were the vitamin D levels in patients that were included in the study?
- Answer: Participants in the study all had less than 30ng/ml of vitamin D, which is below the minimum recommendation of vitamin D levels in the body.

- Original: "Short IBD Questionnaire (SIBDQ) for quality of life"
- Simplification: "a survey about participant's quality of life"
- Rationale: the specific type of survey (SIBDQ) is not mentioned in the simplification
- Question: What survey was used to measure participants' quality of life?
- Answer: The Short Inflammatory Bowel Disease Questionnaire (SIBDQ) was used which gives insights about the physical, social, and emotional status of patients with bowel diseases.
\end{lstlisting}
\end{tcolorbox}
\begin{tcolorbox}[colback=pastelorange!40, colframe=pastelorange, arc=1mm, boxrule=1pt, left=2mm, right=2mm, top=0mm, bottom=0mm]
\begin{lstlisting}
## Original
{{ original }}

## Simplification
{{ simplification }}
\end{lstlisting}
\end{tcolorbox}
\caption{End-to-end system prompt for GPT-4. Colors indicate chat roles: \hlgreen{system} and \hlorange{user}.}
\label{fig:gpt4-prompt}
\end{figure*}

\begin{figure*}
\scriptsize
\begin{tcolorbox}[colback=pastelorange!40, colframe=pastelorange, arc=1mm, boxrule=1pt, left=2mm, right=2mm, top=0mm, bottom=0mm]
\begin{lstlisting}
You are an expert annotator for outputs of text simplification systems. This annotation task is to identify pieces of information that were lost in the simplification process. You will be given two texts: the original and a simplification. Assume that a lay reader only sees the simplification. Identify all things which the reader can NOT learn from the simplification but that they could have learnt from the original.

Please classify each unit of information into one of the following two categories:
1. Omissions: Pieces of information which were not included in the simplification.
2. Imprecise/fuzzy concepts: Pieces of information which are included in the simplification, but that have been simplified to the extent that they became imprecise or completely lost their original meaning.

Afterwards, please write a question-answer pair that would elicit the missing information from the original text. Phrase the question in such a way that a reader can understand it without having seen the original text. It should only contain concepts (entities, events, or states) that were mentioned in the simple text, or concepts that have not been directly mentioned but are generally known or inferable from mentioned ones. The answer should be understandable by an average adult, so please explain technical jargon if necessary. Make each question-answer pair as specific as possible. Ask only about one information unit at a time.

A useful heuristic to decide between omissions and imprecise/fuzzy concepts is to see if the question-answer pair clarifies/expands some topic which is discussed in the simplification. If so, classify it as imprecise/fuzzy concepts, otherwise omission.

Adhere to this output format:
- Original Fragment: <exact span in original>
- Simplification Fragment: null for Omissions OR <exact span in simplification> for Imprecise/fuzzy concepts
- Rationale: <a short rationale that describes what makes this an information loss>
- Question: <the question that elicits missing information>
- Answer: <the answer that provides the missing information from the original span>

Strictly include the above elements. There can be many omissions and imprecise concepts, so be exhaustive. Use simple language in the answer. Cite the original/simplification span EXACTLY so that span == text[text.index(span):text.index(span)+len(span)].

## Original
This study evaluates the effects of vitamin D3 on disease activity and quality of life in ulcerative colitis (UC) patients with hypovitaminosis D. The study was a prospective double-blinded, randomized trial conducted at Community Regional Medical Center, Fresno, CA from 2012-2013. Patients with UC and a serum 25(OH)D level <30 ng/ml were eligible for the study. Enrolled subjects were randomized to receive either 2,000 lU or 4,000 IU of oral vitamin D3 daily for a total of 90 days. The Short IBD Questionnaire (SIBDQ) for quality of life, the Partial Mayo Score for UC disease activity and serum lab tests were compared between the two treatment groups.

## Simplification
This study looks at whether taking vitamin D3 can affect a particular form of bowel disease (ulcerative colitis) and improve the lives of patients with low levels of vitamin D. The study was carried out at a medical center in Fresno, California, between 2012 and 2013. Patients with this disease and low vitamin D levels were included. Participants were randomly given either 2,000 IU or 4,000 IU of oral vitamin D3 daily for 90 days. Researchers used a survey about participant's quality of life and conducted lab tests between the two groups.
\end{lstlisting}
\end{tcolorbox}
\begin{tcolorbox}[colback=pastelblue!40, colframe=pastelblue, arc=1mm, boxrule=1pt, left=2mm, right=2mm, top=0mm, bottom=0mm]
\begin{lstlisting}
## Omissions
- Original Fragment: "Partial Mayo Score for UC disease activity"
- Simplification Fragment: null
- Rationale: the simplification does not mention that the Partial Mayo Score was used.
- Question: Besides quality of life survey and the serum tests, what other measures did the researchers use?
- Answer: The researchers also tracked the partial mayo for UC diseases. UC stands for ulcerative colitis and is an inflammatory bowel disease. The mayo score indicates how active or severe the disease is.

## Imprecise/fuzzy concepts
- Original Fragment: "The study was a prospective double-blinded, randomized trial"
- Simplification Fragment: "The study"
- Rationale: The simplification does not explain the design of the study, it only mentions that it was a "study".
- Question: How did the study control for bias?
- Answer: The study was double-blinded, so that neither the researcher nor the participants knew which treatment each participant received, and it was randomized, meaning the participants were randomly assigned one of the treatments.

- Original Fragment: "a serum 25(OH)D level <30 ng/ml"
- Simplification Fragment: "low vitamin D levels"
- Rationale: the simplification does not explain the inclusion criteria, namely how low the vitamin D levels of eligible patients were.
- Question: How low were the vitamin D levels in patients that were included in the study?
- Answer: Participants in the study all had less than 30ng/ml of vitamin D, which is below the minimum recommendation of vitamin D levels in the body.

- Original Fragment: "Short IBD Questionnaire (SIBDQ) for quality of life"
- Simplification Fragment: "a survey about participant's quality of life"
- Rationale: the specific type of survey (SIBDQ) is not mentioned in the simplification
- Question: What survey was used to measure participants' quality of life?
- Answer: The Short Inflammatory Bowel Disease Questionnaire (SIBDQ) was used which gives insights about the physical, social, and emotional status of patients with bowel diseases.
\end{lstlisting}
\end{tcolorbox}
\begin{tcolorbox}[colback=pastelorange!40, colframe=pastelorange, arc=1mm, boxrule=1pt, left=2mm, right=2mm, top=0mm, bottom=0mm]
\begin{lstlisting}
## Original
{{ original }}

## Simplification
{{ simplification }}
\end{lstlisting}
\end{tcolorbox}
\caption{End-to-end system prompt for Mistral. Colors indicate chat roles: \hlorange{user} and \hlblue{assistant}.}
\label{fig:mistral-prompt}
\end{figure*}

\begin{figure*}
\tiny
\begin{tcolorbox}[colback=pastelgreen!40, colframe=pastelgreen, arc=1mm, boxrule=1pt, left=2mm, right=2mm, top=0mm, bottom=0mm]
\begin{lstlisting}
You are an expert annotator for outputs of text simplification systems. You are given an example, a sample to annotate and detailed instructions.
\end{lstlisting}
\end{tcolorbox}
\begin{tcolorbox}[colback=pastelorange!40, colframe=pastelorange, arc=1mm, boxrule=1pt, left=2mm, right=2mm, top=0mm, bottom=0mm]
\begin{lstlisting}
### Original
This study evaluates the effects of vitamin D3 on disease activity and quality of life in ulcerative colitis (UC) patients with hypovitaminosis D. The study was a prospective double-blinded, randomized trial conducted at Community Regional Medical Center, Fresno, CA from 2012-2013. Patients with UC and a serum 25(OH)D level <30 ng/ml were eligible for the study. Enrolled subjects were randomized to receive either 2,000 lU or 4,000 IU of oral vitamin D3 daily for a total of 90 days. The Short IBD Questionnaire (SIBDQ) for quality of life, the Partial Mayo Score for UC disease activity and serum lab tests were compared between the two treatment groups.

### Simplification
This study looks at whether taking vitamin D3 can affect a particular form of bowel disease (ulcerative colitis) and improve the lives of patients with low levels of vitamin D. The study was carried out at a medical center in Fresno, California, between 2012 and 2013. Patients with this disease and low vitamin D levels were included. Participants were randomly given either 2,000 IU or 4,000 IU of oral vitamin D3 daily for 90 days. Researchers used a survey about participant's quality of life and conducted lab tests between the two groups.

### Missing information in simplification
\end{lstlisting}
\end{tcolorbox}
\begin{tcolorbox}[colback=pastelblue!40, colframe=pastelblue, arc=1mm, boxrule=1pt, left=2mm, right=2mm, top=0mm, bottom=0mm]
\begin{lstlisting}
[
    {
        "category": "imprecise/fuzzy concepts",
        "original": "The study was a prospective double-blinded, randomized trial",
        "simplification": "The study",
        "rationale": "The simplification does not explain the design of the study, it only mentions that it was a 'study'.",
        "question": "How did the study control for bias?",
        "answer": "The study was double-blinded, so that neither the researcher nor the participants knew which treatment each participant received, and it was randomized, meaning the participants were randomly assigned one of the treatments.",
    },
    {
        "category": "imprecise/fuzzy concepts",
        "original": "a serum 25(OH)D level <30 ng/ml",
        "simplification": "low vitamin D levels",
        "rationale": "The simplification does not explain the inclusion criteria, namely how low the vitamin D levels of eligible patients were.",
        "question": "How low were the vitamin D levels in patients that were included in the study?",
        "answer": "Participants in the study all had less than 30ng/ml of vitamin D, which is below the minimum recommendation of vitamin D levels in the body."
    },
    {
        "category": "imprecise/fuzzy concepts",
        "original": "Short IBD Questionnaire (SIBDQ) for quality of life",
        "simplification": "a survey about participant's quality of life",
        "rationale": "The specific type of survey (SIBDQ) is not mentioned in the simplification",
        "question": "What survey was used to measure participants' quality of life?",
        "answer": "The Short Inflammatory Bowel Disease Questionnaire (SIBDQ) was used which gives insights about the physical, social, and emotional status of patients with bowel diseases.",
    },
    {
        "category": "omission",
        "original": "Partial Mayo Score for UC disease activity",
        "simplification": null,
        "rationale": "The simplification does not mention that the Partial Mayo Score was used.",
        "question": "Besides quality of life survey and the serum tests, what other measures did the researchers use?",
        "answer": "The researchers also tracked the partial mayo for UC diseases. UC stands for ulcerative colitis and is an inflammatory bowel disease. The mayo score indicates how active or severe the disease is.",
    },
]
\end{lstlisting}
\end{tcolorbox}
\begin{tcolorbox}[colback=pastelorange!40, colframe=pastelorange, arc=1mm, boxrule=1pt, left=2mm, right=2mm, top=0mm, bottom=0mm]
\begin{lstlisting}
### Original
{{ original }}

### Simplification
{{ simplification }}

### Instructions
This annotation task is to identify pieces of information that were lost in the simplification process. Annotate the sample above. Assume that a lay reader only sees the simplification. Identify all things which the reader can NOT learn from the simplification but that they could have learnt from the original.

Please classify each unit of information into one of the following two categories:
1. Omissions: Pieces of information which were not included in the simplification.
2. Imprecise/fuzzy concepts: Pieces of information which are included in the simplification, but that have been simplified to the extent that they became imprecise or completely lost their original meaning.

Afterwards, please write a question-answer pair that would elicit the missing information from the original text. Phrase the question in such a way that a reader can understand it without having seen the original text. It should only contain concepts (entities, events, or states) that were mentioned in the simple text, or concepts that have not been directly mentioned but are generally known or inferable from mentioned ones. The answer should be understandable by an average adult, so please explain technical jargon if necessary. Make each question-answer pair as specific as possible and make sure that they are independent of each other. Ask only about one information unit at a time.

A useful heuristic to decide between omissions and imprecise/fuzzy concepts is to see if the question-answer pair clarifies/expands some topic which is discussed in the simplification. If so, classify it as imprecise/fuzzy concepts, otherwise omission.

Include following elements in your annotation:
- Original: <exact span in original>
- Simplification: <exact span in simplification> or null for omissions
- Rationale: <a short rationale that describes what makes this an information loss>
- Question: <the question that elicits missing information>
- Answer: <the answer that provides the missing information from the original span>

Strictly follow the above json format. There can be many omissions and imprecise concepts, so be exhaustive. Use simple language in the answer. Cite the original/simplification span EXACTLY so that span == text[text.index(span):text.index(span)+len(span)]. Output ONLY the json!

### Missing information in simplification
\end{lstlisting}
\end{tcolorbox}
\caption{End-to-end system prompt for Llama2. Colors indicate chat roles: \hlgreen{system}, \hlorange{user} and \hlblue{assistant}.}
\label{fig:llama2-prompt}
\end{figure*}

\clearpage
\newpage
\section{Annotation Guidelines}
\label{sec:appendix-annotation-guidelines}
\paragraph{Introduction.} Text simplification aims to rewrite a complex text into a simpler version that can be understood by a lay audience. When simplifying, editors decide what and how to simplify, often omitting content which is deemed less important or too technical. However, this can deny readers access to potentially useful information and the opportunity to learn new terms and concepts.

\paragraph{Goal.} We aim to identify instances where information is lost and to recover it through Question-Answer (QA) pairs. We hypothesize that readers can better understand the text by referring to the simplified version and the associated QA pairs.

\paragraph{The data.} We work with abstracts of randomized controlled trials (RCTs) along with their automatically generated simplifications. RCTs are scientific experiments testing the efficacy of clinical interventions like new drugs, treatments, or diagnostic methods. They typically involve recruiting patients and dividing them into an experimental group (receiving the intervention) and a control group (not receiving it).

\paragraph{Annotation task.} Your task involves the following steps:
\begin{enumerate}[noitemsep]
\item Read both the original and simplified text
\item Compare the two texts and highlight
    \begin{enumerate}[noitemsep]
    \item \hlred{Deletions} from the original, and
    \item \hlorange{Oversimplifications} in the simplification
    \end{enumerate}
\item For each highlight, write a QA pair that re-introduces the information in lay language.
\end{enumerate}
You can find two examples below.

\paragraph{\hlred{Deletions.}} Pieces of information which were not included in the simplification. The questions should allow readers to reveal the omitted content.

\begin{quote}
\small
\underline{Original:} First, nine healthy young men performed two \ding{192} \hlred{20-min} arm-cranking trials \ding{193} \hlred{at 50\% VO2-max} with and without EMS applied to the lower limbs.

\underline{Simplified:} First, they had nine healthy young men do the arm-cranking exercise with and without the EMS added to the lower body.

\emph{Instance 1:}\\
\textbf{Q:} For how long is the exercise applied?\\
\textbf{A:} Participants do the exercise twice for 20 minutes.

\emph{Instance 2:}\\
\textbf{Q:} At what intensity is the exercise applied?\\
\textbf{A:} The exercise is applied at 50\% VO2-max.
\end{quote}

\paragraph{\hlorange{Oversimplifications.}} Pieces of information that were simplified to the extent that they became too imprecise or completely lost their original meaning. The questions should clarify those concepts and restore their original meaning. For this category we highlight both the simplification and the original text that elicited the simplification.

\begin{quote}
\small
\underline{Original:} The \hlorange{flow-mediated vasodilation (FMD)} of the right brachial artery was measured using a high-resolution ultrasound device.

\underline{Simplified:} They used a high-resolution ultrasound to measure \hlorange{how well the main artery in the arm was working.}

\textbf{Q:} How do researchers measure how well the intervention (EMS) works?\\
\textbf{A:} The researchers measure to what extent the main artery of the arm widens, which is called flow-mediated vasodilation (FMD).
\end{quote}

\paragraph{Scenario grounding: deciding what is important enough.}
\begin{itemize}[noitemsep]
\item Do role play. Imagine you are someone who could benefit from understanding the RCT. For example, a patient who has a condition that this RCT addresses.
\item When in doubt, try to take an inclusive perspective. People can always decide a question is not relevant to them.
\item For numerical results like p-values or Z-scores, make a judgment if they are necessary for a correct understanding of the RCT. Do not interpret these results, rather rely on the authors interpretation of the values.
\end{itemize}

\paragraph{Tips and other notes.}
\begin{itemize}[noitemsep]
\item When is something a deletion vs. oversimplification? A useful heuristic is to see if you could ``attach'' the omitted information somewhere to the simplification. If so, it is likely an oversimplification.
\item You are free to use a search engine for writing the answers. Please only use high quality sources.
\item The text may refer to the same concept multiple times. Please highlight all occurrences, and copy-paste the QA.
Add any comments, questions or concerns in the comment field.
\item Ignore other artifacts in the simplifications that are not about an information loss, including factual errors, under-simplifications (i.e., when something is still too complicated) and unnecessary information (e.g., the registration number of a trial).
\end{itemize}

\paragraph{Checklist for a good QA pair.}

\begin{itemize}[noitemsep]
\item Address an information gap between the original and the simplification.
\item A question should be self-contained. Readers should get a sense of “why” it is an important question to ask and “what” they will learn if they look at it.
\begin{itemize}[noitemsep]
    \item Deletions: keep in mind that readers only see the simple text. So you may need to add a bit more context into the question to make it apparent to readers “why” this is important.
    \item Oversimplifications: explicitly connect to the concept that is being clarified. This can often be done by somehow including the concept in the question.
\end{itemize}
\item Question scoping: try to phrase the question such that the highlighted piece is the most likely answer, ideally there should be a singular answer.
\item The answer must draw on information from the original.
\item The answer must use plain language that an average adult would understand. You may have to explain difficult terms and concepts.
\item For each QA pair, please check that the answer cannot be found at a later stage in the simplification. The reason is that a simplification may reorder content or drop redundant information to reduce the length. We define information loss globally, so be aware of these cases. Basically, this is a small sanity check that you can do while drafting the QA pair.
\item \textbf{Avoid:} Yes/no questions (rather, reformulate it using a question word: what, how, why, etc.).
\item \textbf{Avoid:} Questions that ask for further simplification or external information (these are not about an information gap).
\end{itemize}

\begin{figure}[b!]
\centering
\includegraphics[width=.98\textwidth]{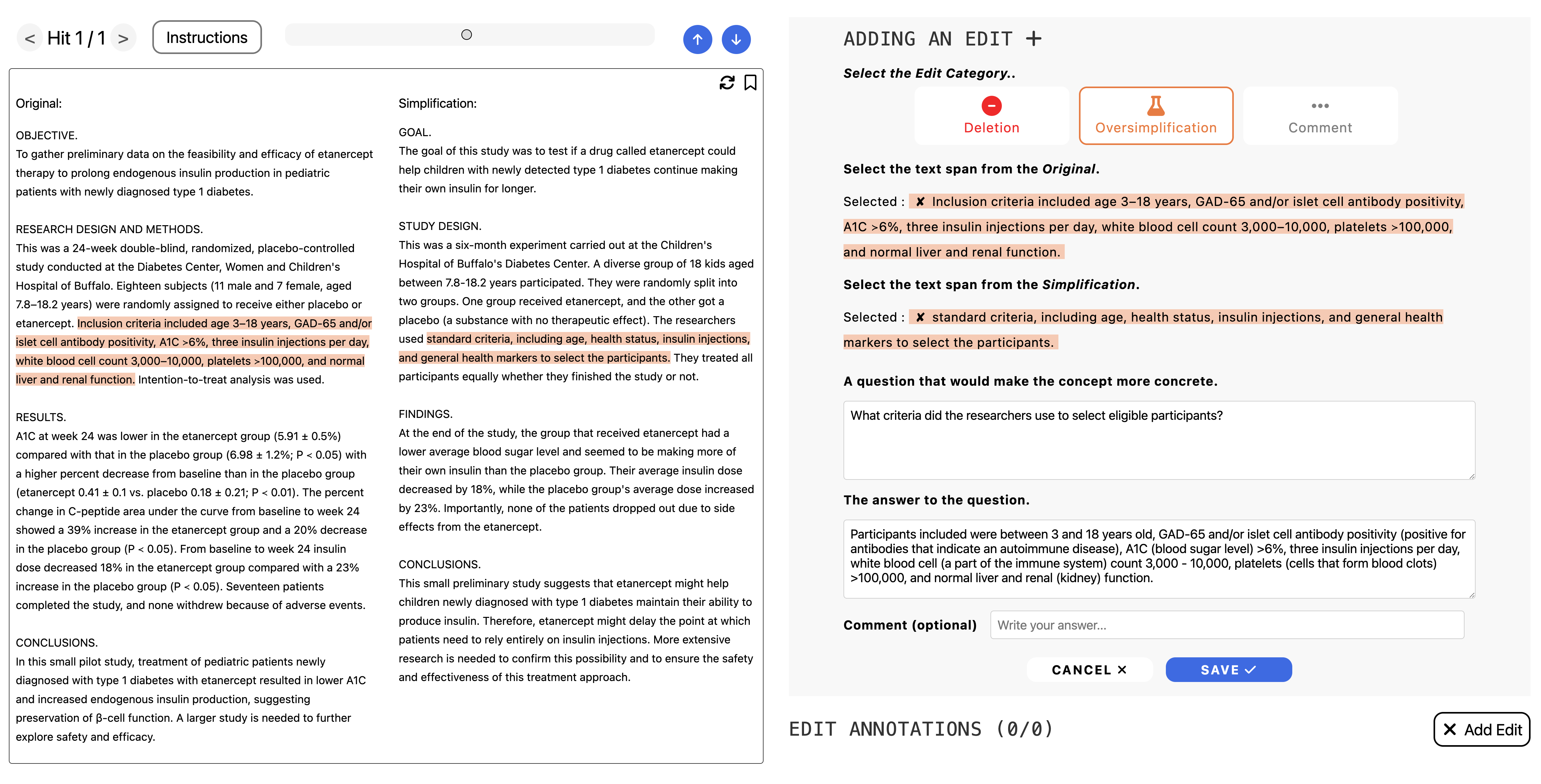}
\caption{Annotation interface for \textsc{InfoLossQA} built with Thresh~\cite{Heineman:2023:EMNLP}.}
\label{fig:annotation-interface}
\end{figure}

\newpage
\section{Evaluation Guidelines}
\label{sec:appendix-evaluation-guidelines}
We seek to evaluate models that identify information loss caused by simplifying text. These models provide two outputs: (i) a localization of what information was lost, and (ii) a QA pair that elicits the missing information.

\subsection{Model Recall of Human-written QA}
We aim to determine how many of the human-written QAs are generated by the models. This is a pairwise comparison. You will be given a reference QA and a model QA. Please assign one of the following categories:
\begin{itemize}[noitemsep]
\item \hlgreen{Fully recalled:} the model QA gives the same information as the reference QA
\item \hlorange{Partially recalled:} the model QA partially answers the reference QA
\item \hlred{Not recalled:} there is no overlap in the presented content
\end{itemize}
As our goal is to identify if models identify the same information loss, we should abstract away from surface level dissimilarities and framing of questions. Here are some guidelines to help with this judgment:

\begin{itemize}[noitemsep]
\item First, identify the specific unit of information that the reference is asking about by looking both at the reference question, answer and (if needed) the localization. Then check if this information is conveyed by the model answer. The full texts and answer localizations are given for your reference. Note, however, that the model localization may not reflect the generated QA.
\item A useful heuristic is to check how well the reference answer could be replaced by the model-generated answer, even if this may lead to a slightly weird/unintuitive phrasing.
\item Disregard background explanations. When a reference answer explains a technical concept but the model answer does not include this explanation, it can still be fully recalled. We evaluate simplicity separately.
\item Both the reference or the model-generated QA can include more information that cannot be matched to the other QA. For example, the model QA could ask a broader question which entails multiple reference QAs.
\item In examples where both QAs are asking for the same information, but the model answer misinterprets the information, we mark as not recalled.
\end{itemize}
See the table below for example annotations.

\begingroup
\renewcommand{\arraystretch}{1.3}
\small
\begin{longtable}{lp{0.3\textwidth}p{0.3\textwidth}lp{0.15\textwidth}}
\toprule
\textbf{\#} & \textbf{Reference QA} & \textbf{Model QA} & \textbf{Rating} & \textbf{Comment} \\
\midrule
\endfirsthead
\multicolumn{5}{@{}l}{\ldots continued}\\
\toprule
\textbf{\#} & \textbf{Reference QA} & \textbf{Model QA} & \textbf{Rating} & \textbf{Comment} \\
\midrule
\endhead 
\bottomrule
\multicolumn{5}{r@{}}{continued \ldots}\\
\endfoot
\bottomrule
\endlastfoot
1
& \textbf{How were the carbon nanoparticles administered to the participants?}
This study was a dose-escalation study, meaning that the carbon nanoparticles were administered in increasing doses sequentially to different groups of participants.
& \textbf{What type of study was conducted to test the effects of carbon nanoparticles?}
An investigator-initiated, randomized controlled, dose-escalation study was conducted.
& \hlgreen{Full}
& Disregard background explanation in the original.\\

2
& \textbf{What condition did the babies in the study have?} The babies in the study were diagnosed with moderate-to-severe bronchiolitis.
& \textbf{How was the study conducted?} The study was a double-blind, randomized controlled trial on infants (1 to 12 months) who were diagnosed in the emergency department with moderate-to-severe bronchiolitis.
& \hlgreen{Full}
& Fully recalled even if phrased differently\\

3
& \textbf{What statistical tests were used to analyze the results?} The results were analyzed using a repeated measures (the procedures were conducted multiple times), mixed design ANOVA test (compares the results from three or more study groups).
& \textbf{What type of statistical analysis was used to analyze the results?} A repeated measures, mixed design ANOVA was used to analyze the results.
& \hlgreen{Full}
& Disregard background explanation\\

4
& \textbf{What was done to prevent bias in the trial?} The trial was designed as a double-blind trial, meaning neither the patients nor the researchers knew which group of participants had been assigned which combination of sedative drugs.
& \textbf{How was this study conducted?} This study was conducted as a clinical trial at a single center. It was double-blinded, meaning that neither the researcher nor the participants knew which treatment each participant received, and it was randomized, meaning the participants were randomly assigned one of the treatments.
& \hlgreen{Full}
& Model QA is quite broad but includes the missing bit of information (double-blind).\\

5
& \textbf{What was the major criterion for patients to participate in the study?} The 155 patients who participated in this study had a glaucoma-related diagnosis, meaning they had been diagnosed with an eye condition that causes vision loss.
& \textbf{What kind of health condition did all enrolled patients have?} All enrolled participants were diagnosed with a glaucoma-related condition.
& \hlgreen{Full}
& Main subject is ``major criterion'' which the model QA identifies. 155 patients is not the subject of the question, and background explanation ``vision loss'' can be ignored.\\

6
& \textbf{How were the carbon nanoparticles administered to the participants?} This study was a dose-escalation study, meaning that the carbon nanoparticles were administered in increasing doses sequentially to different groups of participants.
& \textbf{What were the dosage amounts of the carbon nanoparticles used in the study?} Carbon nanoparticles were given in doses of \SI{10}{\micro\gram}, \SI{50}{\micro\gram}, or \SI{100}{\micro\gram}.
& \hlorange{Partial}
& The question asks for dose-escalation. The answer shows increasing doses, but it misses calling out ``dose escalation.''\\

7
& \textbf{What specific parts of the digestive system were observed?} The study looked at the antrum (part of the upper stomach) and the second duodenal portion (lower part of the stomach).
& \textbf{In this study, where were the muscle contractions assessed?} The muscle contractions were assessed in the antrum, which is a part of the stomach.
& \hlorange{Partial}
& \\

8
& \textbf{What other analyses were done of the rsFC strength?} Correlation analyses were done to see if the rsFC strength under heroin exposure was connected to the rewarding effect and plasma (in the blood) concentrations of heroin.
& \textbf{Did the researchers also look for a relationship between the strength of the brain connections and the level of morphine in the patients' blood?} Yes, the researchers studied whether there was a correlation between the strength of resting state functional connectivity (rsFC) and the plasma concentration of morphine, a major metabolite of heroin.
& \hlorange{Partial}
& Rewarding effect is missing.\\

9
& \textbf{What kind of tests were used to assess patients in this study?} Participants were assessed with laboratory tests, the United Kingdom screening test [...], the Michigan neuropathy screening score [...], and the Michigan diabetic neuropathy score [...]. In addition, [...]
& \textbf{What method was used to assess the level of neuropathy in the patients?} The Michigan neuropathy screening score was used to assess the level of neuropathy in all patients.
& \hlorange{Partial}
& One model generates relatively narrow questions which can touch upon a small part of a larger reference question.\\

10
& \textbf{What kind of factors is the study analyzing to determine how effective the fast track area (FTA) method is?} The study is analyzing the impact of a fast tract area (FTA) using both effectiveness measures, such as waiting times and length of stay in the waiting room, and quality measures, such as mortality rates and LWBS rates (patients who leave without being seen by a medical professional).
& \textbf{What is the purpose of the fast track area?} The purpose of the fast track area is to reduce waiting times, patient dissatisfaction, and morbidity by providing a separate area for less urgent cases, allowing for faster treatment and better patient flow.
& \hlred{No}
& Superficially a lot of overlap, but the answer here needs to make clear \emph{what is measured} not what the fast track area can hypothetically do.\\

11
& \textbf{What aspects of heart function were increased using the combination of LS and SNP drugs?} Left ventricular systolic (the heart's ability to pump blood) and diastolic (heart relaxation and how long it takes for the ventricles, two chambers of the heart, to fill with blood) functions including LVEF (how much blood is pumped out of the heart each time it contracts (beats)) improved using the combination of LS and SNP drugs.
& \textbf{What was the main conclusion of the study?} The study shows that the LS and SNP combination really improved heart function during and after surgery. Using these drugs led to less need for other supporting drugs or equipment. Thus, we suggest using this drug combination before and during surgery.
& \hlred{No}
& Does not give an answer about the actual heart function. \\

12
& \textbf{How much did the special questionnaire overestimate calcium intake compared to the 24-hour recall?} The questionnaire overestimated the average total calcium intake by 221 mg/d (milligrams per day) compared to the 24-hour recall across racial groups.
& \textbf{What was the variability in daily calcium intake estimated by the 24-hour recall?} The variability or standard deviation in daily calcium intake estimated based on the 24-hour dietary recall was 433 milligrams per day.
& \hlred{No}
& \\

13
& \textbf{How did the other treatments perform in comparison to each other?} The results showed that Group 2 that received only the laser treatment showed better results than Group 1 who received only a low-dose of potassium nitrate.
& \textbf{What were the specifics of the treatment groups in the study?} The treatment groups in the study were: Group 1 - 5\% potassium nitrate, Group 2 - gallium-aluminum-arsenide diode laser with a wavelength of 980 nm, noncontact pulse mode, and power wattage of 0.5 W, Group 3 - combined 5\% KNO3 and the diode laser, and Group 4 - placebo (control).
& \hlred{No}
& \\
\end{longtable}
\endgroup

\begin{figure}[hb]
\centering
\includegraphics[width=.98\textwidth]{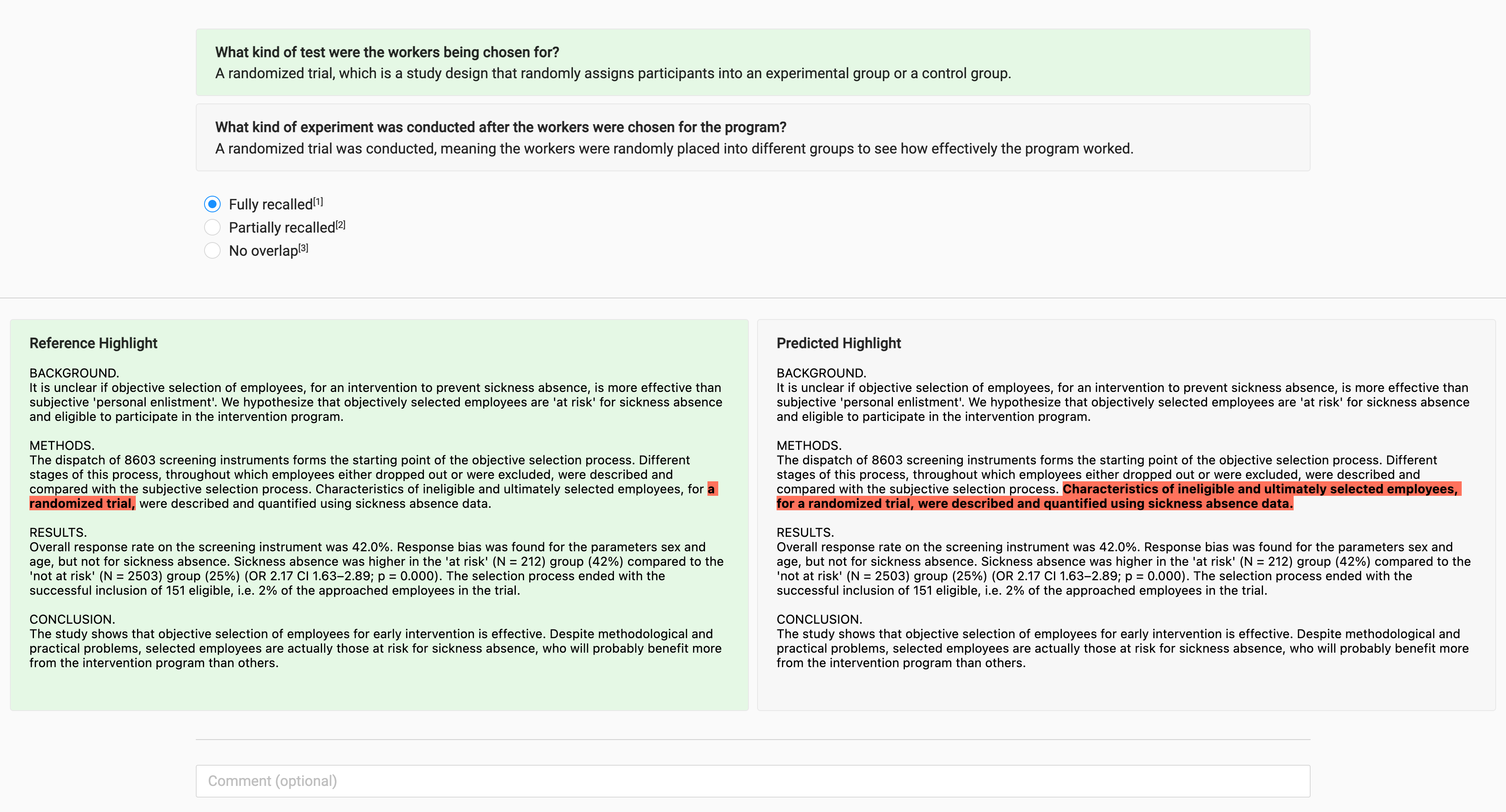}
\caption{Interface for pairwise evaluation of QA recall. Built with \rurl{LabelStud.io}.}
\label{fig:evaluation-interface-recall}
\end{figure}

\clearpage

\subsection{Quality of Generated QA}

\paragraph{Motivation/rationale (all questions).} For each rating, please provide a brief explanation that motivates your choice. For positive ratings, the rationale can be omitted. For negative ratings, explain how the QA pair could be improved to get to a positive rating. This helps us to analyze the answers and to improve the annotation guidelines.

\paragraph{Criterion 1: question givenness.} A question should be interpretable for a reader. It should only contain concepts (entities, events, or states) that were mentioned in the question context or concepts that are generally known or inferable from mentioned ones. We define question context as follows: For deletions, the context is the entire simple text. For oversimplifications, the context is everything in the simple text up to and including the question localization. For this evaluation, please pretend that you only see the simplified text.

\vspace{0.5em}
\noindent \emph{How is the question phrased?}
\begin{todolist}[noitemsep,topsep=0pt]
\item Good (reader focused, no new concepts)
\item Bad (e.g., question introduces new concepts, answer leakage, hallucinations)
\end{todolist}

\paragraph{Criterion 2: question localization.}
When the question seeks more information/clarification about an oversimplified concept, the corresponding span in the simplified text should be highlighted. For ``Missing:'' highlight the corresponding text in the simple text which discusses the topic in an oversimplified way.

\vspace{0.5em}
\noindent \emph{To what extent does the highlight relate to the topic under discussion?}
\begin{todolist}[noitemsep,topsep=0pt]
\item Good: the highlight corresponds to the topic that the question discusses
\item Unrelated: the highlighted text does not relate to what the question is asking
\item Missing: there should be a highlight, but there is not (please add highlight...)
\item n/a: the topic under discussion is not part of the simple text (= deletion)
\end{todolist}

\paragraph{Criterion 3: answer simplicity.}
The answer should be easy to understand. Please focus on the readability and simplicity of the answer. This is different from accuracy which we will evaluate later.

\vspace{0.5em}
\noindent \emph{Does the answer contain jargon?}
\begin{todolist}[noitemsep,topsep=0pt]
\item The answer is jargon-free
\item The answer contains jargon but it is adequately explained in the answer
\item The answer contains jargon but it is adequately explained in the simplified text
\item The answer contains unexplained jargon
\end{todolist}

\vspace{0.5em}
\noindent \emph{Is the answer standalone?}
\begin{todolist}[noitemsep,topsep=0pt]
\item Yes, the answer can be understood without looking at the original
\item No, the answer contains confusing aspects (e.g., unresolved coreferences, abbreviations/acronyms)
\end{todolist}

\paragraph{Criterion 4: answerability/question Relevance.}
The question should be about an information loss between the original and simplified texts. We evaluate this in two steps: answerability on the original text, and answerability on the simplified text. A question is about an information loss if it is answerable on the original, but unanswerable/only vaguely answerable on the simplified text.

When making your assessment, you may use the answer snippet to help with this evaluation. However, be aware that the snippet may not correctly answer the question. In those cases, disregard the snippet and look for other places which answer the question.

\vspace{0.5em}
\noindent \emph{Is the question answerable with the \ul{original text}?}
\begin{todolist}[noitemsep,topsep=0pt]
\item Yes, and there is a single obvious answer
\item Yes, but there could be multiple valid answers
\item No
\end{todolist}

\vspace{0.5em}
\noindent \emph{To what extent is the question answerable with the \ul{simplified text} (i.e., degree of information loss)?} The benchmark for this is the answer on the original text.
\begin{todolist}[noitemsep,topsep=0pt]
\item Fully answerable. Asking it on the simplified text would give the same answer or a closely paraphrased answer as on the original.
\item Partly or vaguely answerable. The simplified text gives some relevant information, but is less specific or exhaustive than the original.
\item Unanswerable.
\end{todolist}

\paragraph{Criterion 5: answer accuracy.}
The question should be correctly answered. This criterion only applies to questions that can be answered with the original text. Unanswerable questions are out of scope of this investigation. 
Ideally a question should be specific enough so that there is a singular answer (see Criterion 4: answerability). If the question is ambiguous/vague (i.e., there are multiple valid answers), its answer has a high chance of being incomplete or only partially answering the question.

\vspace{0.5em}
\noindent \emph{Does the \ul{provided answer} correctly answer the question?}
\begin{todolist}[noitemsep,topsep=0pt]
\item Yes
\item Partially, the answer is related but misses information
\item No
\end{todolist}

\vspace{0.5em}
\noindent \emph{Does the \ul{provided answer} have any hallucinations?} Hallucinations are information or claims that cannot be traced back to the original. Disregard general background explanations and elaborations.
\begin{todolist}[noitemsep,topsep=0pt]
\item Good: there are no hallucinations
\item Bad: the answer contains hallucinations
\end{todolist}

\vspace{0.5em}
\noindent \emph{Does the \ul{answer snippet} correctly answer the question? Imagine it was rephrased into a fluent answer.}
\begin{todolist}[noitemsep,topsep=0pt]
\item Yes
\item Partially, the answer snippet is related but misses information
\item No
\end{todolist}

\begin{figure}[hb]
\centering
\includegraphics[width=.98\textwidth]{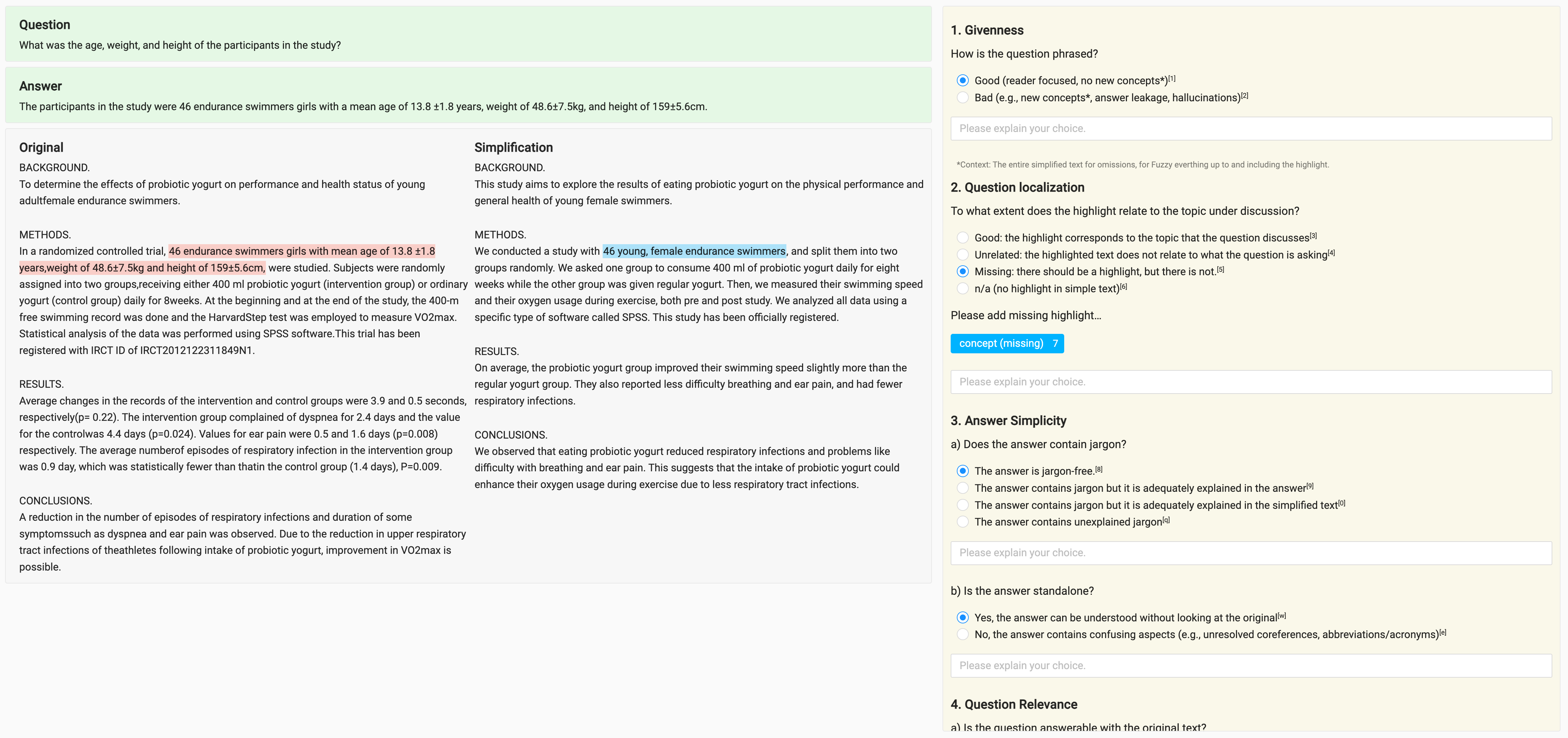}
\caption{Interface for quality assessment of QA (criteria continued in scrollbar). Built with \rurl{LabelStud.io}.}
\label{fig:evaluation-interface-accuracy}
\end{figure}

\section{Data Release and License}
We reused RCT abstracts from the \emph{Evidence Inference V2.0} dataset~(\citealp{DeYoung:2020:BioNLP}; {\small \rurl{evidence-inference.ebm-nlp.com}}, accessed 2024-05-28). After personal communication with the authors, it was confirmed that all articles in this dataset were from the PubMed Open Access subset which only includes license terms that allow reuse ({\small \rurl{ncbi.nlm.nih.gov/pmc/tools/openftlist}}, accessed 2024-05-28). After discussion with our institutions' librarian on fair use, we release the annotations in \textsc{InfoLossQA} under CC-BY-4.0.

\end{document}